\documentclass{article}
\usepackage[top=1in, bottom=1in, left=1in, right=1in]{geometry}
\usepackage{amsmath,amsthm,amssymb,amsfonts,mathtools,bm,bbm,array,graphicx,float,mathdots,dsfont,mathrsfs,latexsym}
\usepackage{caption,subcaption,multirow,longtable,lscape,wrapfig,tikz,tikz-cd,comment}
\usepackage[colorlinks=true, pdfstartview=FitV,linkcolor=blue,citecolor=blue,urlcolor=blue]{hyperref}
\usepackage[toc,page]{appendix}
\usepackage{booktabs}
\usepackage{multirow}
\usepackage{algorithm}
\usepackage{algpseudocode}
\DeclareMathOperator{\Z}{\mathbb{Z}}
\usepackage{blkarray}
\usepackage{amsmath}

\newtheorem{conjecture}{Conjecture}[section]

\renewcommand{\thefootnote}{\fnsymbol{footnote}}

\newcommand{\hflogo}{\raisebox{-0.2em}{\includegraphics[height=1em]{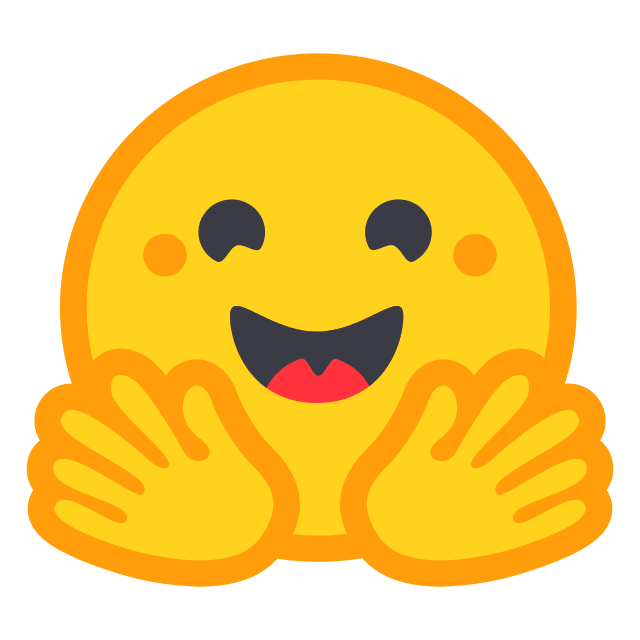}}}
\newcommand{\ghlogo}{\raisebox{-0.2em}{\includegraphics[height=1em]{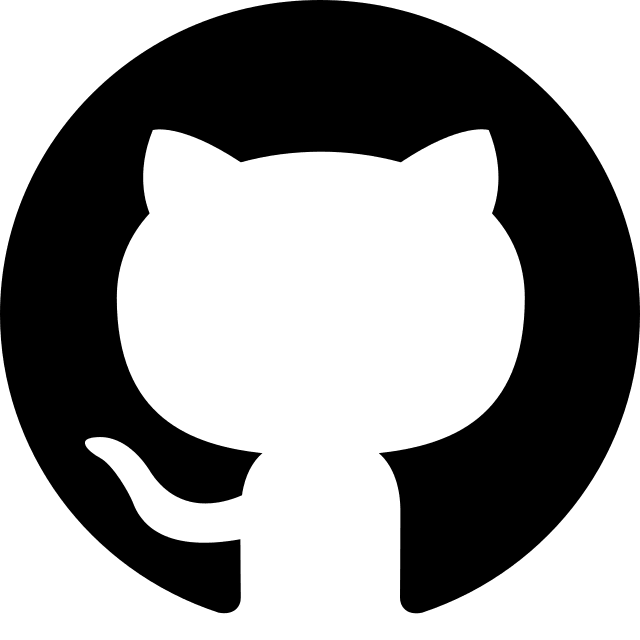}}}

\title{Learning the Graphical Nature of Symmetries}
\author{Rashid Barket\thanks{\texttt{barketr@coventry.ac.uk}} \textsuperscript{1},
Enrico Grimaldi\thanks{\texttt{enrico.grimaldi@uniroma1.it}} \textsuperscript{2},
Yacoub Hendi\thanks{\texttt{yacoub.hendi@math.uu.se}} \textsuperscript{3},
Edward Hirst\thanks{\texttt{ehirst@unicamp.br}} \textsuperscript{4},\\
Adam Onus\thanks{\texttt{adi.onus@ludwig.ox.ac.uk}} \textsuperscript{5,6,7},
Harmeet Singh\thanks{\texttt{harmeet.2.singh@kcl.ac.uk}} \textsuperscript{8}}
\date{\small\today \\
\textsuperscript{1} \textit{Coventry University, Coventry, UK}\\
\textsuperscript{2} \textit{Department of Computer, Control and Management Engineering,
Sapienza University of Rome, Rome, Italy}\\
\textsuperscript{3} \textit{\small Uppsala University, Uppsala, Sweden}\\
\textsuperscript{4} \textit{\small Instituto de Matemática, Estatística e Computação Científica (IMECC) da Universidade Estadual de Campinas (UNICAMP), 13083-859, Brasil}\\
\textsuperscript{5} \textit{\small Ludwig Institute for Cancer Research, Nuffield Department of Medicine, University of Oxford, Oxford, UK}\\
\textsuperscript{6} \textit{\small Tumour Cell Biology Laboratory, The Francis Crick Institute, London, UK }\\
\textsuperscript{7} \textit{\small Mathematical Institute, University of Oxford, Oxford, UK}\\
\textsuperscript{8} \textit{\small King's College London, London, UK}\\
}

\begin{document}
\maketitle
\begingroup
\renewcommand{\thefootnote}{}
\footnotetext{Authors listed alphabetically.}
\endgroup
\renewcommand{\thefootnote}{\arabic{footnote}}
\setcounter{footnote}{0}

\begin{abstract}
Finite groups are rigid algebraic objects, whose Cayley graphs expose a rich network geometry through which group-theoretic structure can be measured, compared, and learned. In this paper, a dataset of $131{,}406$ Cayley graphs is constructed, covering all groups of order at most $767$ except order $512$, recording exact algebraic labels for group properties together with a broad collection of graph, cycle, distance, and spectral statistics. This census aims to provide novel benchmarks for studying how finite-group properties are reflected in Cayley graph observables. It also yields new enumerative contributions: alongside recovering known OEIS sequences for standard group classes, new sequences for monolithic groups and for groups generated by at most three, four, and five elements are contributed to the OEIS. The accompanying network analysis identifies several empirical regularities and formulates testable conjectures, including relationships involving square clustering, Cayley graph diameter, average graph disorder, and spectral eigengaps of nilpotent groups. Finally, a comparison between classical models, an MLP, and graph neural network architectures is performed for predicting algebraic group properties directly from Cayley graph data. The results show that engineered graph statistics are highly informative, while GNNs, especially GIN and in some fixed-order settings GCN, can recover substantial structural signal directly from the graph. Such that graph-aware architectures show phases of optimality on these group-theoretic graph representations.\\
% \textbf{Code:} \href{https://github.com/Engrima18/CayleyNet}{\texttt{github.com/Engrima18/CayleyNet}}
\small Dataset: \hflogo \, \href{https://huggingface.co/collections/Enrico18/cayley-graphs-of-finite-groups}{\texttt{huggingface.co/collections/Enrico18/cayley-graphs-of-finite-groups}}.
\\
\small Code: \ghlogo \, \href{https://github.com/Engrima18/CayleyNet}{\texttt{github.com/Engrima18/CayleyNet}}.
\end{abstract}

\newpage
\numberwithin{equation}{section}
\numberwithin{figure}{section}
\numberwithin{table}{section}
\tableofcontents
\newpage

%%%%%%%%%%%%%%%%%%%%%%%%%%%%%%%%%%%%%%%%%%%%%%%%%
\section{Introduction}
Symmetry is one of the organising principles of modern mathematics and theoretical science, and group theory provides its most flexible algebraic language. Finite groups arise naturally in geometry, number theory, combinatorics, coding theory, and mathematical physics, where one seeks to understand not only the abstract multiplication law of a group but also the structural fingerprints that distinguish one family of symmetries from another. A particularly effective way to externalise this algebraic structure is to pass from a group to one of its Cayley graphs. Given a generating set, the Cayley graph converts multiplication by generators into a graph whose combinatorial geometry encodes algebraic information such as growth, diameter, relations among generators, and the extent to which the group behaves like a highly constrained or highly diffuse symmetry object. Because Cayley graphs are regular and vertex-transitive, they offer a mathematically rigid setting in which global graph statistics can be studied without many of the confounding effects present in general networks.

Data-driven methods have also become standard across the mathematical and physical sciences. In particular, graph neural networks and related representation-learning methods have developed into a mature toolkit for graph-structured data \cite{wu2020comprehensive, xu2018powerful, bouritsas2022improving}. In mathematical settings these methods are especially attractive, since the underlying objects often carry exact symmetries and rigorously defined invariants, allowing one to ask whether computationally accessible features recover genuinely structural information. %Recent work has already shown that machine learning can detect and even help formulate nontrivial statements about finite groups \cite{he2023learningsimple}.

This motivates the present study, which sits at the interface of computational group theory, network analysis, and graph machine learning. Cayley graphs have recently emerged as a natural meeting point for these subjects. They have been used as testbeds for learning group-theoretic properties \cite{he2023learningsimple}, for machine-learning-based pathfinding and diffusion on exponentially large discrete state spaces \cite{douglas2025diffusionmodelscayleygraphs, cayleypy2}, for large-scale experimentation on growth and diameter phenomena \cite{cayleypy3}, and even as a framework in which broader theoretical analogies between discrete geometry and AI are explored \cite{cayleypy4}. On the graph-learning side, Cayley graph structure has already been proposed as a principled mechanism for improving information propagation in graph neural networks \cite{wilson2025cayleygraphpropagation}. To the best of our knowledge, the present work is the first to apply graph neural networks directly to Cayley graphs as the mathematical data objects under study, rather than using Cayley-graph structure only as an auxiliary propagation template. In particular, Cayley graphs are treated here as supervised graph data on which graph-learning models, including GCN and GIN type architectures, are evaluated alongside more classical baselines.

The aim of this paper is to provide such a study. A large dataset of Cayley graphs associated with finite groups from GAP's Small Groups library is constructed \cite{GAP4}, together with group-theoretic labels and graph-theoretic summary statistics. The central question is whether readily computable network features of a Cayley graph encode enough information to predict structural properties of the underlying group, including whether the group is abelian, nilpotent, simple, perfect, solvable, monolithic, or cyclic. From the mathematical side, this probes which algebraic phenomena leave a detectable imprint on coarse graph observables, assessed with network analysis techniques to build novel conjectures. From the machine-learning side, it yields a controlled benchmark with exact labels and intrinsic symmetries. The main contributions are therefore an exhaustive dataset up to fixed order (767) of finite groups containing group-theoretic and network analysis properties; an extensive analysis of correlations with network properties of these graphs leading to novel conjectures; and a comparative study of several machine-learning models on classifying these group-theoretic properties, including graph neural networks applied directly to Cayley graphs. The resulting picture is not that Cayley graphs provide a generator-independent fingerprint of a group, but rather that even for generator-dependent constructions, substantial and interpretable statistical signal survives.

The remainder of the paper is organised as follows. In Section \ref{sec:background} the relevant background on Cayley graphs, graph-based learning, and recent computational work at the interface of group theory and AI is reviewed. Section \ref{sec:dataset_gen} describes the construction of the dataset and the graph-theoretic and algebraic quantities that were recorded. Section \ref{sec:analysis} studies the resulting network statistics and their correlations with group properties. Section \ref{sec:ML} presents the machine-learning experiments and comparative performance of the models considered, with Section \ref{sec:summary} then summarising the main results.

\section{Background}
\label{sec:background}

The basic mathematical object studied throughout is the Cayley graph of a finite group. For a finite group $G$ and a generating set $S \subseteq G$, the directed Cayley graph $\mathrm{Cay}(G,S)$ is defined by
\begin{equation}
\mathrm{Cay}(G,S) = \bigl(G,\, \{(g,gs) : g \in G,\ s \in S\}\bigr).
\end{equation}
When $S$ is inverse-closed, such that $(g,gs)$ and $(gs,g)$ are both in $\mathrm{Cay}(G,S)$, an undirected graph would also be suitable. By construction, every vertex has out-degree (and in-degree) $|S|$, and left multiplication by a fixed group element induces a graph automorphism. Cayley graphs are therefore regular and vertex-transitive, while still permitting meaningful variation in diameter, girth, clustering, cycle structure, and spectral behaviour.

By passing from algebra to graph theory, questions about generators, relations, and word length are recast as questions about paths, cycles, and graph distance. The resulting translation is generator-dependent, since different generating sets can produce markedly different Cayley graphs for the same group. That dependence is nevertheless informative, as it records how a chosen generating set exposes the geometry of the group. In computational settings, it is therefore natural for both algebraic invariants and graph-derived observables to be studied simultaneously.

\paragraph{Graph-theoretic notation.}
It will be convenient throughout to fix some standard notation for general graphs, of which Cayley graphs are a special case. A graph is a pair $H = (V,E)$, with $V$ a finite set of \emph{nodes} and $E \subseteq V \times V$ a set of \emph{edges}; $H$ is \emph{directed} or \emph{undirected} according to whether edges carry orientations, and we write $|V|$ and $|E|$ for the cardinalities. The structure of $H$ is fully encoded by a small set of algebraic operators: the \emph{adjacency matrix} $\mathbf{A} \in \{0,1\}^{|V|\times|V|}$, with $\mathbf{A}[i,j]=1$ iff $(i,j) \in E$; the diagonal \emph{degree matrix} $\mathbf{D} = \mathrm{diag}(\deg(v_1), \dots, \deg(v_{|V|}))$; and the combinatorial \emph{graph Laplacian} $\mathbf{L} = \mathbf{D} - \mathbf{A}$, which for undirected $H$ is symmetric and positive semidefinite, and whose spectral characterisation carries useful interpretations for the corresponding graph. For instance, the Laplacian's smallest eigenvalue $\lambda_1 = 0$ multiplicity is equal to the number of connected components, and the Laplacian eigenvectors can be used for defining clustering \cite{von2007tutorial} or sampling strategies over graphs \cite{tsitsvero2016signals}. These operators indeed underpin spectral methods in graph signal processing \cite{ortega2018graph} and will reappear in the network statistics of Section~\ref{sec:net_stats} and the graph neural network architectures of Section~\ref{sec:ML}. When $S$ inverse-closed, the adjacency matrix of $\mathrm{Cay}(G,S)$ symmetric, every node has degree $|S|$, and the Laplacian inherits the vertex-transitive symmetry of the underlying group.

\paragraph{Graph signals and convolutional filters.}
A useful additional abstraction is that of a \emph{graph signal}: a function $\mathbf{x}: V \to \mathbb{R}$, identified with the vector $\mathbf{x} \in \mathbb{R}^{|V|}$ that assigns a numerical value to each node. Propagation of a signal along the topology is described by a \emph{Graph Shift Operator} (GSO): a linear map $\mathbf{S} \in \mathbb{R}^{|V|\times|V|}$ satisfying $\mathbf{S}[i,j]=0$ whenever $(i,j)\notin E$ and $i \neq j$, so that the shifted signal at any node aggregates the signal values over its closed neighbourhood. Canonical instances of the GSO are $\mathbf{S} = \mathbf{A}$, yielding a one-step diffusion, and $\mathbf{S} = \mathbf{L}$, yielding a local discrepancy. Building on the GSO, a \emph{graph convolutional filter} of order $K$ is defined as the matrix polynomial
\begin{equation}
    \mathrm{\Phi}(\mathbf{S}) \;=\; \sum_{k=0}^{K} \varphi_k \, \mathbf{S}^k,
    \label{eq:graph-filter}
\end{equation}
with learnable coefficients $\varphi_0,\dots,\varphi_K$ weighting the contributions of successive $k$-hop neighbourhoods. The resulting class of operators is linear, shift-invariant, and permutation-equivariant, providing the principled inductive bias that underlies the graph neural network models used in Section~\ref{sec:ML} \cite{ortega2018graph}.

\paragraph{Cayley graphs in graph machine learning.}
This perspective is closely aligned with modern graph-based machine learning. Graph neural networks and related architectures have become standard tools for learning on relational data \cite{wu2020comprehensive}; their expressive power has been related to the Weisfeiler--Leman hierarchy \cite{xu2018powerful}, and subgraph-count augmentation has been shown to improve graph-level expressivity \cite{bouritsas2022improving}. For mathematical datasets, these results suggest that global and mesoscopic graph statistics may carry substantial information beyond raw adjacency data. Cayley graphs have also been adopted as computational templates within graph learning. In Cayley Graph Propagation, full Cayley graph structure is used to mitigate over-squashing more effectively than truncated expander-based constructions \cite{wilson2025cayleygraphpropagation}. The present work is different in emphasis: to the best of our knowledge, it is the first to use graph neural networks directly on Cayley graphs themselves as labelled mathematical graph data, rather than using Cayley-graph structure only to improve message propagation on other learning tasks. On the algebraic side, machine learning has already been applied directly to finite-group data, where neural networks were used to study simplicity among generated subgroups of symmetric groups, and a conjecture suggested by the model was subsequently proved \cite{he2023learningsimple}. Algorithmic problems on Cayley graphs have likewise attracted recent attention. Pathfinding on Cayley graphs and group actions was formulated within a diffusion-model framework in \cite{douglas2025diffusionmodelscayleygraphs}, while in the CayleyPy RL project, reinforcement learning and diffusion-distance methods were combined for large-scale pathfinding experiments \cite{cayleypy2}. Within the broader CayleyPy programme, a library for Cayley and Schreier graph computation was introduced, large-scale experiments on growth and diameter were reported, and numerous conjectures motivated by those data were proposed \cite{cayleypy3}. In a further instalment, more speculative links between learning tasks on Cayley graphs and holographic ideas from mathematical physics were explored \cite{cayleypy4}.

The present paper is complementary to this literature. Rather than focusing on a single family of groups or on a single algorithmic task, a broad census of finite groups is studied in order to identify how standard network observables of Cayley graphs relate to classical algebraic properties. Emphasis is placed on dataset construction, descriptive network analysis, and comparative prediction, so that both a benchmark dataset and an empirical baseline for subsequent work at the interface of algebra, networks, and machine learning are provided.

%%%%%%%%%%%%%%%%%%%%%%%%%%%%%%%%%%%%%%%%%%%%%%%%%
\section{Dataset Generation}
\label{sec:dataset_gen}

In order to generate a large dataset of Cayley graphs of groups, we used the computer algebra software GAP \cite{GAP4} which itself contains a database of all groups of order up to $2000$ (excluding those of order $1024$), the so called \emph{small groups library}.\footnote{In fact, for certain orders, information is stored for groups of order beyond $2000$, see \url{https://docs.gap-system.org/pkg/smallgrp/doc/chap1.html} for details.} While GAP itself does not have an in-built function that returns the Cayley graph of a group, using the networkx package \cite{networkx_docs} we can build the Cayley graph from groups in GAP's small groups library via the following algorithm.

\begin{algorithm}
\caption{Construct Cayley Graph}
\begin{algorithmic}[1]
\Require A group $G$
\Ensure A directed graph representing the Cayley graph of $G$

\State $\text{elements} \gets \text{Elements}(G)$
\State $\text{generators} \gets \text{GeneratorsOfGroup}(G)$
\State $\text{graph} \gets$ empty directed graph

\ForAll{$g \in \text{elements}$}
    \State Add node labeled $g$ to $\text{graph}$
\EndFor

\ForAll{$g \in \text{elements}$}
    \ForAll{$s \in \text{generators}$}
        \State $h \gets g \cdot s$
        \State Add directed edge from $g$ to $h$ with label $s$
    \EndFor
\EndFor

\State \Return $\text{graph}$
\end{algorithmic}
\end{algorithm}

The Cayley graph is then stored as a networkx directed graph. In order to start with a generating set for our group, we used the in-built GAP function \verb|GeneratorsOfGroup|. We note that this is not guaranteed to give a minimal generating set for our group (though in some instances it does). For example, in GAP, the cyclic group of order $6$ is \verb|SmallGroup(6,2)| and running \verb|GeneratorsOfGroup| on \verb|SmallGroup(6,2)| returns two generators despite only one generator being needed (this comes from the fact that the cyclic group of order $6$ can be decomposed as $\Z/6\Z \simeq \Z/2\Z \times \Z/3\Z$). In terms of Cayley graphs, having extra generators amounts to multiple edges between the same node (so our graphs are not simple in general). We note that GAP does have the in-built function \verb|MinimalGeneratingSet| which returns a generating set of minimal size, but this function is often excessively computationally intensive for groups of large order.

\begin{figure}
    \centering
    \includegraphics[width=0.5\linewidth]{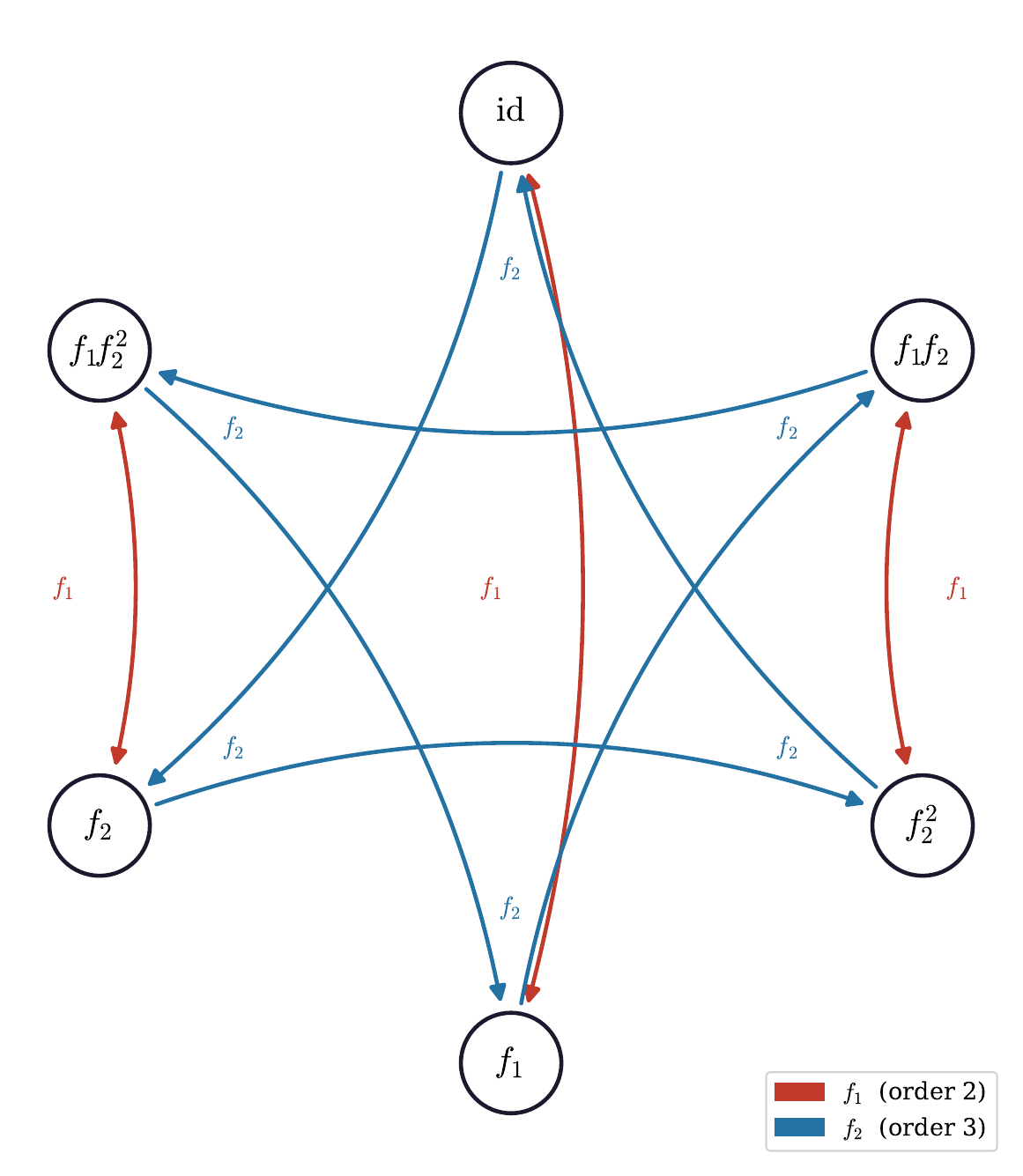}
    \caption{Cayley Graph visualisation \cite{hansen2011visualization} of the group $S_{3}$ with generators $\{f_{1}, f_{2}\}$ where $f_{1}$ has order $2$ and $f_{3}$ has order $3$.}
    \label{fig:s3_group}
\end{figure}

In order to encode the Cayley graph in our dataset, we then take the adjacency matrix of the Cayley graph and record the indices of the non-zero entries. For example, for the group with GAP Id $6,1$ (which is the symmetric group $S_{3}$), the above algorithm gives the Cayley graph reported in Figure~\ref{fig:s3_group}.
The associated adjacency matrix is then 
\[
A = \begin{blockarray}{ccccccc}
& id & f_{1} & f_{2} & f_{1}f_{2} & f_{2}^{2} & f_{1}f_{2}^{2} \\
\begin{block}{c(cccccc)}
id & 0 & 1 & 1 & 0 & 0 & 0\\
f_{1} & 1 & 0 & 0 & 1 & 0 & 0\\
f_{2} & 0 & 0 & 0 & 0 & 1 & 1\\
f_{1}f_{2} & 0 & 0 & 0 & 0 & 1 & 1\\
f_{2}^{2} & 1 & 0 & 0 & 1 & 0 & 0\\
f_{1}f_{2}^{2} & 0 & 1 & 1 & 0 & 0 & 0\\
\end{block}
\end{blockarray}
 \]
which we record in our dataset as the list 
\[
[[0, 1], [0, 2], [1, 0], [1, 3], [2, 4], [2, 5], [3, 4], [3, 5], [4, 0], [4, 3], [5, 1], [5, 2]].
\]
The advantage of recording only the nonzero entries is in saving storage (recall that the adjacency matrix for a Cayley graph will be a square matrix with dimension the order of the group). With the Cayley graph generated for each group, we then used the following GAP functions to compute various group theoretic properties that form the target variable in our machine learning experiments (see \cite{Dummit1999AbstractA} for more details on the various definitions):
\begin{enumerate}
    \item[] \verb|IsAbelian|: is the group operation commutative?
    \item[] \verb|IsNilpotent|: does the group have an ascending upper central series which terminates in the whole group?
    \item[] \verb|IsSimple|: are the only normal subgroups the entire group and the trivial subgroup?
    \item[] \verb|IsPerfect|: is the group equal to its commutator subgroup?
    \item[] \verb|IsSolvable|: does the group have a descending series of commutator subgroups which terminates in the trivial subgroup?
    \item[] \verb|IsMonolithic|\footnote{Note this function did not exist in GAP and was coded by our team.}: is there a unique nontrivial minimal normal subgroup?
    \item[] \verb|IsCyclic|: is the group generated by a single element?
\end{enumerate}

In addition to these target variables, we also generated other features, namely recording the minimal number of generators needed for each group (to offset the possibility that \verb|GeneratorsOfGroup| does not always return a minimal generating set) as well as the following network properties computed from the generated Cayley graph using the networkx package (see \cite{networkx_docs} for details):
\begin{itemize}
    \item The number of triangles out of each node in the Cayley graph (this is independent of the node due to node-transitivity of Cayley graphs).
    \item The density and girth of the Cayley graph.
    \item The average clustering coefficient and square clustering coefficients of each Cayley graph.
    \item The Wiener, Schultz and Gutman index of the Cayley graphs.
    \item The list of the cycle lengths and their frequency present in a minimal cycle basis of the Cayley graph.
    \item The diameter of the Cayley graph computed both via an approximation giving a lower bound (using \verb|nx.algorithms.approximation.diameter|) and also by an exact computation (using the function \verb|nx.algorithms.distance_measures.diameter|), complementing the conjectures and work in \cite{cayleypy3}.
\end{itemize}

In order to keep track of which generating elements were used for which edge of the Cayley graph, we also recorded a one hot encoded vector signalling the generator used between each pair of connected nodes. We note that some of the network properties above are not valid on directed graphs, in which case we adopted the convention that if the network property can be computed on directed graphs, we compute it on the directed graph, otherwise we compute it on the underlying undirected graph. We also began computing the eigenvector centrality of each Cayley graph but realised this simply returned the reciprocal of the square root of the order of the group (since Cayley graphs are regular so $(1,\dots,1)$ is always an eigenvector). In order to generate our data, we used a high performance computing (HPC) cluster accessed through King's College London \cite{HPC_Cluster}. In doing so, we were able to generate a complete dataset recording all the above mentioned properties and Cayley graphs for all groups of order up to $767$, excluding the groups\footnote{Note that there are $10,494,213$ groups of order $512$.} of order $512$. In total this gives us a dataset with $131,406$ Cayley graphs. The reason for stopping at $767$ was due to the large number of groups of order $768$ ($1,090,235$ groups in total). The full code and data scripts used are available in the accompanying \href{https://github.com/Engrima18/CayleyNet}{\texttt{GitHub}}\footnote{\href{https://github.com/Engrima18/CayleyNet}{\texttt{https://github.com/Engrima18/CayleyNet}}} repository.

%%%%%%%%%%%%%%%%%%%%%%%%%%%%%%%%%%%%%%%%%%%%%%%%%
\section{Data Analysis}\label{sec:analysis}
Before building our machine learning architecture, we first performed extensive analysis on our data.
This was to identify any non-trivial correlations within and between the network statistics and graph properties which may be of independent interest for novel conjectures, in addition to informing on what input data to add to our model.

\subsection{Group statistics}
\begin{figure}
    \centering
    \includegraphics[width=0.8\linewidth]{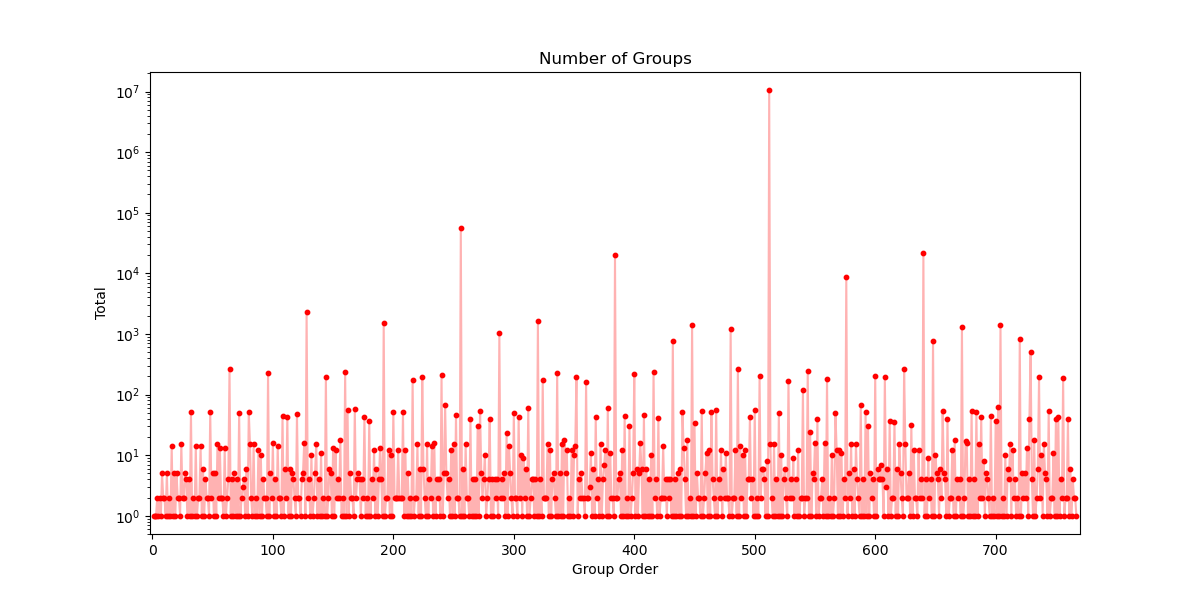}
    \caption{The number of groups of each order up to 767.}
    \label{fig:num_groups}
\end{figure}

The number of groups of each order up to 767 is shown in Figure~\ref{fig:num_groups}, and is consistent with sequence A000001 in The On-line Encyclopedia of Integer Sequences (OEIS) \cite{oeis}.
Approximately $40\%$ (56,092 in total) of the groups in our dataset are of order 256, which we treat separately for most of our analysis.
There are also a large number of groups of order 384 (20,169 groups), 576 (8,681 groups) and 640 (21,541 groups).
In general, orders with a large number of prime factors have a large number of groups, which is related to the theory of group extensions; $256=2^8$, $384=3\times2^7$, $512=2^9$, $576=3^2\times 2^6$ and $640=5\times 2^7$.

\begin{figure}
    \centering
    \includegraphics[width=0.49\linewidth]{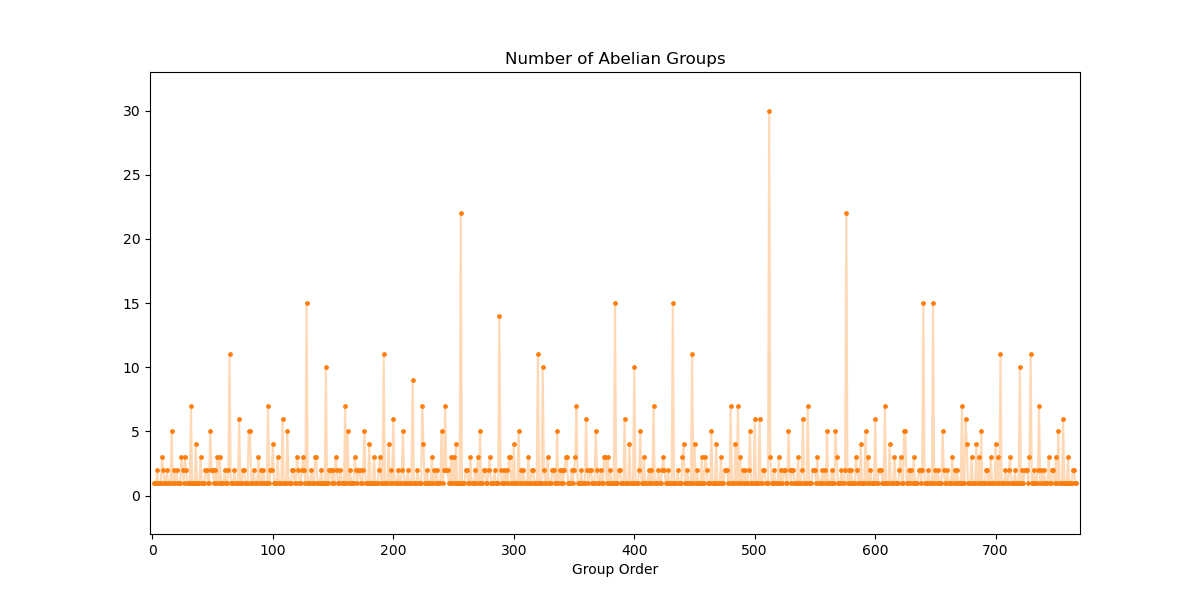}
    \includegraphics[width=0.49\linewidth]{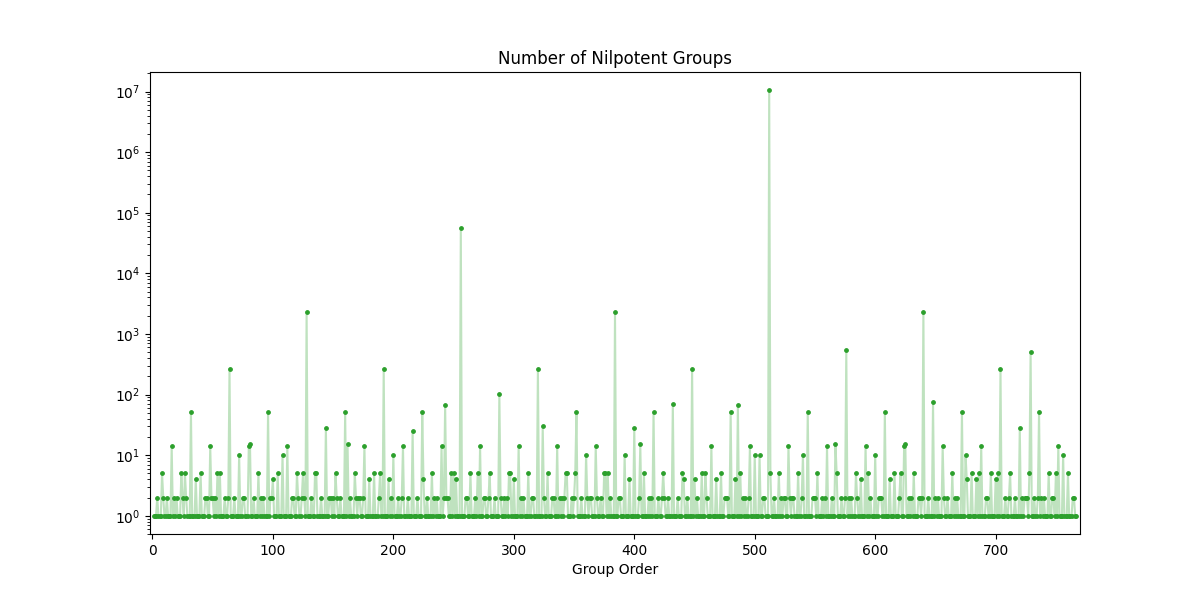}
    \includegraphics[width=0.49\linewidth]{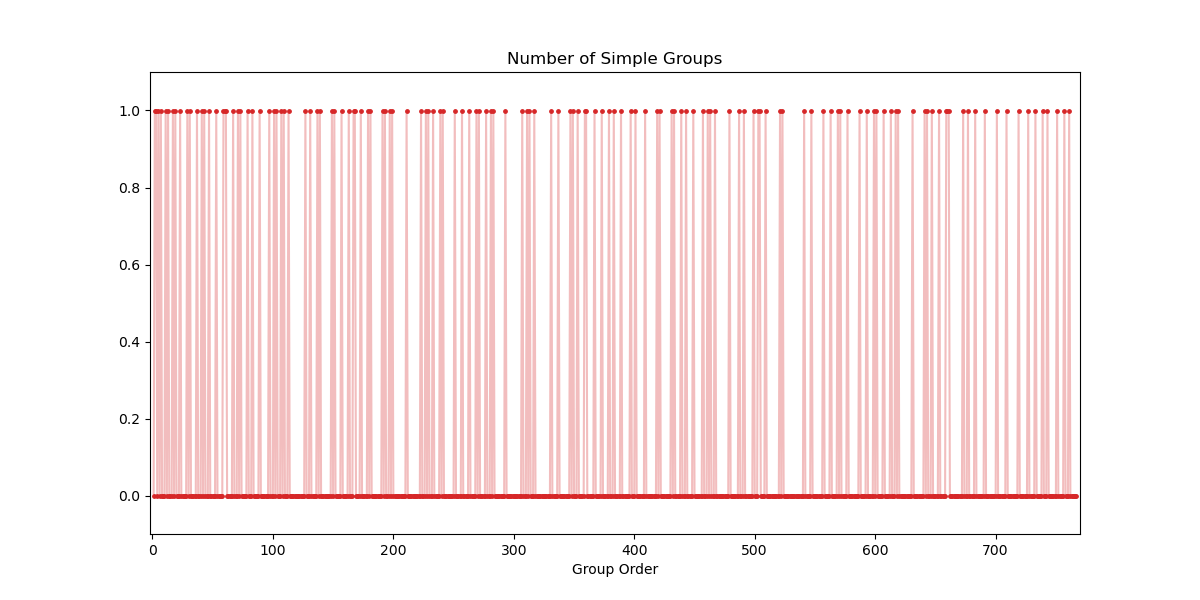}
    \includegraphics[width=0.49\linewidth]{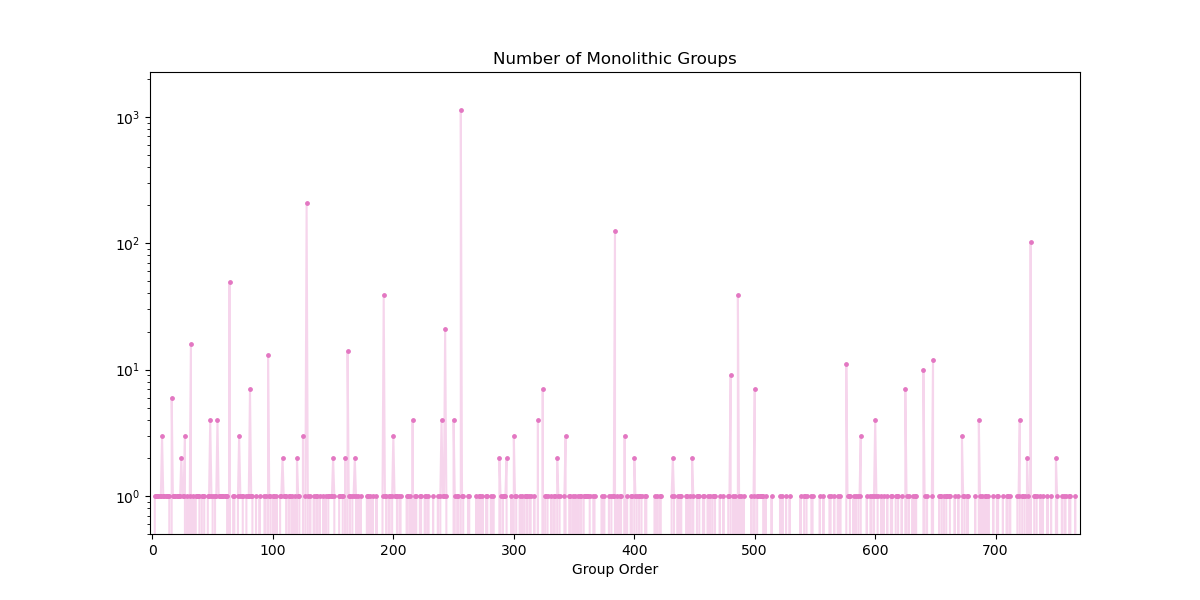}
    \caption{The number of abelian (top-left), nilpotent (top-right), simple (bottom-left) and monolithic (bottom-right) groups of each order up to 767. Note that each plot has a different scale and we do not have any date for monolithic groups of order 512.}
    \label{fig:group_properties}
\end{figure}

We also record whether each group is cyclic, abelian, nilpotent, simple, perfect, solvable or monolithic, and all our results agree with known group theory.
To summarise some of these properties: 
\begin{itemize}
    \item All cyclic groups are abelian.
    \item All cyclic groups of order a prime power are simple.
    \item All abelian groups are nilpotent and all nilpotent groups are solvable.
    \item All groups of order a prime power are nilpotent.
    \item All simple groups are monolithic.
    \item All non-cyclic simple groups are perfect.
    \item All perfect groups are non-solvable.
\end{itemize}
We also record the number of groups with each group property, and present examples in Figure~\ref{fig:group_properties}.
To summarise:
\begin{itemize}
    \item There is exactly one cyclic group for every order.
    \item The number of abelian groups agrees with sequence A000688 of the OEIS.
    \item The number of nilpotent groups agrees with sequence A066060 of the OEIS.
    \item The number of simple groups agrees with sequence A005180 of the OEIS.
    \item The number of perfect groups agrees with sequence A060793 of the OEIS.
    \item The number of solvable groups agrees with sequence A056866 of the OEIS (when subtracted from sequence A000001).
    \item We were unable to find any previous reference with which to compare the number of monolithic groups, which we have now \textit{published} in the OEIS as the \textit{new} sequence A395682.
\end{itemize}

\begin{figure}
    \centering
    \includegraphics[width=0.8\linewidth]{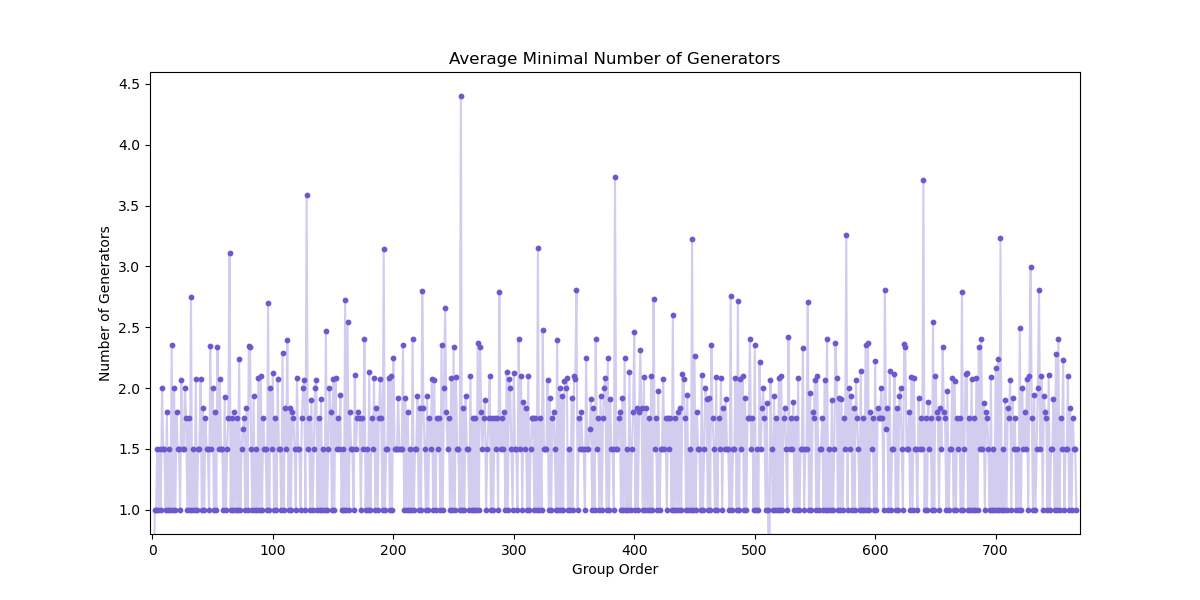}\\
    \includegraphics[width=0.8\linewidth]{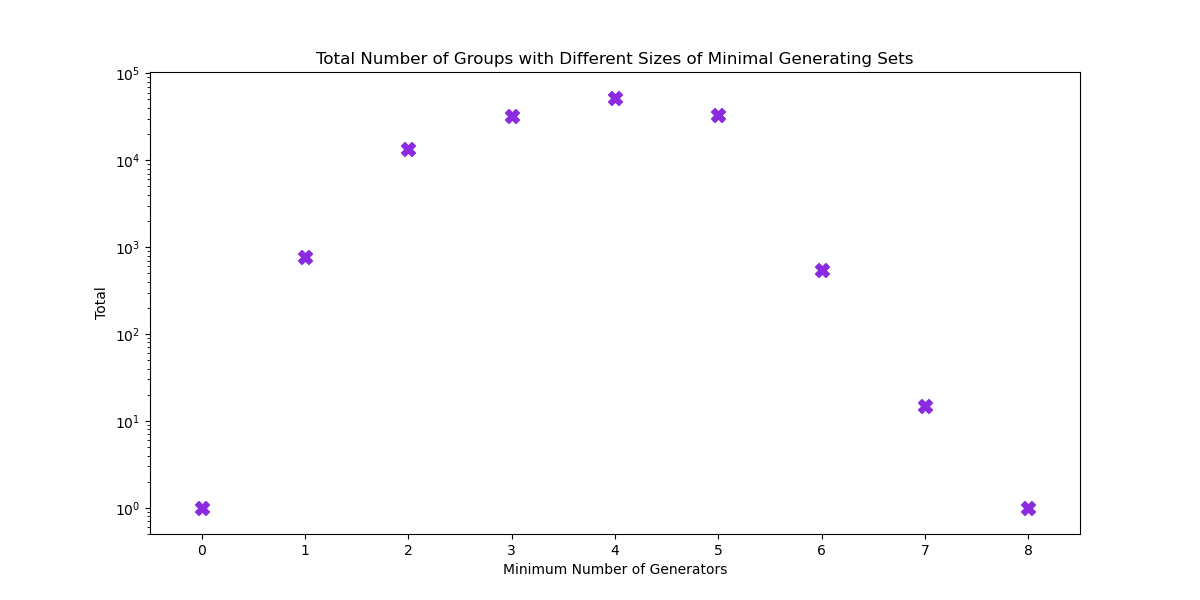}\\
    \includegraphics[width=0.8\linewidth]{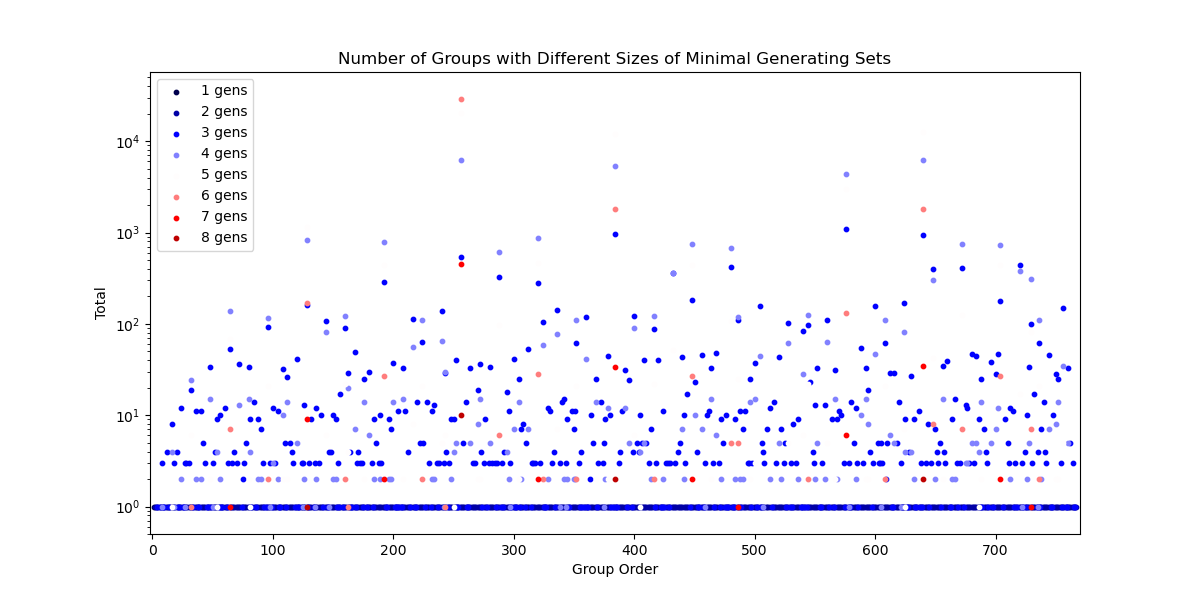}
    \caption{A summary of size of minimum generating sets for groups in our dataset. Top: the mean number of minimum generators for groups of each order. Middle: the total number of groups with different sizes of minimum generating sets. Bottom: distribution of number of groups of each order with fixed minimum number of generators, with minimum number of generators varying by colour.}
    \label{fig:min_gens}
\end{figure}

We also recorded the minimum number of generators for each of our groups, a summary of which can be found in Figure~\ref{fig:min_gens}.
The number of groups with at most two minimal generators agrees with sequence A066389 of the OEIS, but we were unable to find any previous reference with which to compare groups of three or more generators.
We have \textit{published} the number of groups with at most three, four and five minimal generators in the OEIS as sequences A396721, A394617 and A394618 respectively.
In these sequences, we observed that there is exactly one group of $2^n$ with a minimal generating group of size $n$ -- namely $(\Z/2\Z)^n$ -- and all other groups of order $2^n$ can be generated with fewer than $n$ elements.
This a consequence of known group theory (in particular Burnside's Basis Theorem which relates minimal generating sets and Frattini subgroups) and can be generalised: for any prime $p$, the only group of order $p^n$ which requires at least $n$ generators is $(\Z/p\Z)^n$.

\subsection{Network Statistics: Definitions and Interpretations}
\label{sec:net_stats}
Here we outline the explicit formulas for graph invariants\footnote{These invariants have already been informative in other network science analyses for mathematics \cite{Costantino:2024joa, Dechant:2022ccf}.} which we compute with the networkx package and interpret them directly as a function of the Cayley graph presentation of a group where appropriate.
In what follows, let $G$ be a group of order $n$ with (not necessarily minimal) generating set $S$ where $|S|=g$ and let $H=(V,E)$ be a (possibly directed) graph, with distance function $d(\cdot,\cdot)$ between nodes (of the underlying undirected graph).

\paragraph{Density and Diameter:}
The density $\mathcal{D}$ and the diameter $\mathfrak{D}$ of a graph are measures of the connectivity of a graph.
Specifically, the density measures the proportion of (directed) edges compared to the maximal number of edges (i.e. for a complete directed graph) and the diameter measures the largest distance between nodes (of an undirected graph);
\begin{align}
\mathcal{D}(H) & = \frac{|E|}{|V|(|V|-1)} \label{eq:density} \\
\mathfrak{D}(H) & = \max_{u,v\in V}d(u,v).
\end{align}
As $\mathrm{Cay}(G,S)$ is a regular graph where all nodes have in and out degree $g$ (and therefore all nodes in the underlying directed graph have degree $2g$), there are $ng$ edges.
This means $\mathcal{D}(\mathrm{Cay}(G,S))=g/(n-1)$, which provides no additional information about Cayley graphs which could not have been discerned immediately from the group presentation.
On the other hand, $\mathfrak{D}(\mathrm{Cay}(G,S))$ represents the largest non-redundant representation of all elements of $S$, therefore implicitly encoding non-trivial information about the group relations.

\paragraph{Topological Indices:}
The Wiener index ($\mathcal{W}$) is the sum of (undirected) distances between all pairs of nodes, and the Gutman and Schultz\footnote{The Schultz index is sometimes called the \textit{degree distance index.}} indices ($\mathcal{G}$ and $\mathcal{S}$) are variants which weight this sum by node degrees.
%These indices are commonly used in chemical graph theory \cite{rouvray2002topology,ashraf2022wiener} and are defined as follows
\begin{align}
\mathcal{W}(H) & = \frac{1}{2}\sum_{u\in V}\sum_{v\in V}d(u,v) = \sum_{u\neq v \in V}d(u,v) \label{eq:wiener}\\
\mathcal{G}(H) & = \sum_{u\neq v \in V}d(u,v)\deg(u)\deg(v) \label{eq:gutman}\\
\mathcal{S}(H) & = \sum_{u\neq v \in V}d(u,v)\left[\deg(u)+\deg(v)\right] \label{eq:schultz}
\end{align}
Related to these indices is also the average graph disorder ($\mathcal{A}$) which normalises $\mathcal{W}(G)$ by the number of nodes: $\mathcal{A}(G)=2\mathcal{W}(G)/n$.
While this is not directly in our data or our experiments, we still included this in our analysis for completeness.

For $\mathrm{Cay}(G,S)$, the underlying undirected graph which we use to compute distances and degrees has degree $2g-g_2$ at each node, where $g_2$ is the number of generators of $S$ of order 2 ($s$, so that $gss=g$).
Here we need to remove $g_2$ because each incoming and outgoing edge of the unweighted graph will be identified.
Thus we get $\mathcal{G}(\mathrm{Cay}(G,S))=(2g-g_2)^2\,\mathcal{W}(\mathrm{Cay}(G,S))$ and $\mathcal{S}(\mathrm{Cay}(G,S))=(4g-2g_2)\,\mathcal{W}(\mathrm{Cay}(G,S))$.
In this special setting, the sum of distances from a single fixed node to all other nodes is constant, so
\[
\mathcal{A}(\mathrm{Cay}(G,S)) = \frac{2}{n}\mathcal{W}(\mathrm{Cay}(G,S)) = \frac{2}{n}\frac{1}{2}\sum_{g\in G}\sum_{h\in G}d(g,h) = \sum_{h\in G}d(g^*,h).
\]
where $g^*$ is a fixed element of $G$ (say, the identity).
Thus, the average disorder is only a function of the generating set $S$ and (implicitly) the relations of $G$.

\paragraph{Cycles:}
The girth ($\mathfrak{G}$) of a (directed) graph is the minimum (directed) cycle length.
This is related to the number of triangles ($\mathcal{T}$) in $H$; $\mathfrak{G}(H)=3$ if and only if $\mathcal{T}(H)>0$.
In the case of $\mathrm{Cay}(G,S)$, the number of triangles at each node (say, $\mathcal{T}_{node}$) is constant and $\mathcal{T}(\mathrm{Cay}(G,S))=n\,\mathcal{T}_{node}(\mathrm{Cay}(G,S))/3$.
Both $\mathcal{T}_{node}(\mathrm{Cay}(G,S))$ and $\mathfrak{G}(\mathrm{Cay}(G,S))$ encode information about the group relations, and form a subset of the information contained in the minimal cycle basis, which we also compute for each datapoint.

In networkx, the minimal cycle basis is calculated for undirected graphs and is recorded as two lists; a list of cycle lengths and a corresponding list of the number of independent cycles of these lengths.
For numerical analysis, we then recorded the maximum cycle length, the sum of lengths of all cycles and the average cycle length of the basis for each group.
We also computed the size of the basis, although this stat can be immediately derived from the Euler characteristic ($\chi$) of the graph; the number of cycles ($\beta_1$) of a connected (undirected) graph satisfies 
\[
\beta_1(H) = \chi(H)+1 := |E|-|V|+1.
\]
Recall that each node of the underlying Cayley graph has degree $2g-g_2$, so there are $n(g-\frac{1}{2}g_2)$ edges in total\footnote{This is always an integer, since there can only be generators of order two if $n$ is even.} and $\beta_1(\mathrm{Cay}(G))=(g-\frac{1}{2}g_2-1)n+1$.
As with the graph density, this does not provide additional information not already available to us.

\paragraph{Clustering coefficients:} The average clustering coefficient ($\mathcal{C}$) measures how clustered a node in the network is on average, and is closely related to the number of triangles passing through a node $(\mathcal{T}_v$ for node $v$).
For a (directed) graph, the clustering coefficient around each node is the ratio of the number of (directed) triangles compared to the total amount possible which could be formed (as determined by the degree of the node).
In our dataset, we compute this for directed graphs, which is
\begin{equation}
\mathcal{C}(H) = \frac{1}{n}\sum_{v\in V}\frac{\mathcal{T}_v(H)}{2\left[\deg(v)\left(\deg(v)-1\right)-2\deg_{\leftrightarrow}(v)\right]},
\label{eq:cluster_coeff}
\end{equation}
where $\deg(v)=\deg_{\mathrm{in}}(v)+\deg_{\mathrm{out}}(v)$ and $\deg_{\leftrightarrow}(v)$ is the number of directed cycles of length two.
For $\mathrm{Cay}(G,S)$, $\deg_{\leftrightarrow}(g)$ counts the number of generators of order two and twice the number of pairs of generators which are inverses ($s$ and $t$ such that $gst=gts=g$).
In our dataset, no generating sets contain pairs of inverses, so if $g_2$ counts the number of generators in $S$ of order two, we have
\[
\mathcal{C}(\mathrm{Cay}(G,S)) = \frac{1}{n}\sum_{v\in V}\frac{\mathcal{T}(\mathrm{Cay}(G,S))}{2\left[2g(2g-1)-2g_2\right])}=\frac{\mathcal{T}(\mathrm{Cay}(G,S))}{4(2g^2-g-g_2)}
\]
which provides exactly the same information as $\mathcal{T}(\mathrm{Cay}(G))$ when the generating set is known.

Similarly, we may define the (average) square clustering coefficient ($\square$) as the ratio of the number of (directed) cycles of length four (squares) compared to the total amount which could possibly be formed.
In our dataset, we compute this for directed graphs, which is
\begin{equation}
\square(H) = \frac{1}{n}\sum_{v\in V}\frac{\sum_{u\in N_\mathrm{out}(v)}\sum_{w\in N_\mathrm{in}(v)\setminus u}\left(\#\left[N_\mathrm{out}(u)\cap N_\mathrm{in}(w)\right]-1\right)}{\sum_{u\in N_\mathrm{out}(v)}\sum_{w\in N_\mathrm{in}(v)\setminus u}\left(\deg_\mathrm{out}(u)+\deg_\mathrm{in}(w)-\#\left[N_\mathrm{out}(u)\cap N_\mathrm{in}(w)\right]-2\theta_{uw}-1\right)}
\label{eq:square_cluster}
\end{equation}
where $N_\mathrm{in}(v)$ and $N_\mathrm{out}(v)$ is the set of neighbours with edges in and out of $v$ (respectively), $\#$ denotes the size of a set so that $\deg_\mathrm{in}(v)=\# N_\mathrm{in}(v)$ and $\deg_\mathrm{out}(v)=\# N_\mathrm{out}(v)$, and $\theta_{uw}$ is the $(u,w)$ entry of the graph adjacency matrix ($1$ if there is an edge from $u$ to $w$ and $0$ otherwise).
Since $\mathrm{Cay}(G,S)$ is regular, we need only compute the fractional term for a fixed node (and not take any summation), otherwise it is difficult to simplify the expression without prior knowledge of the presentation (even for the undirected counterpart).

\subsection{Network Statistics: Correlations and Trends}
We now outline the following trends and correlations we observed in the data.
In the interest of brevity, we do not report on trivial correlations which are obvious from the group properties mentioned above and/or by the definition the network statistics mentioned above.
Moreover, while we find some interesting trends and are able provide conjectures from data analysis alone, we were not able to find any definitive relationships between network statistics and group properties, which shows that any positive result in group identification from machine learning is non-trivial.

\begin{figure}
    \centering
    \includegraphics[width=0.4\linewidth]{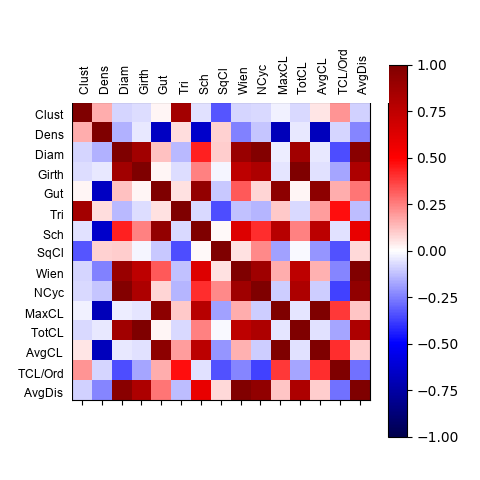}
    \includegraphics[width=0.4\linewidth]{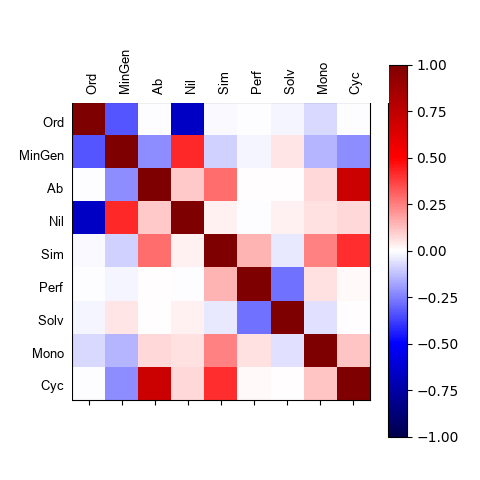}
    \includegraphics[width=0.6\linewidth]{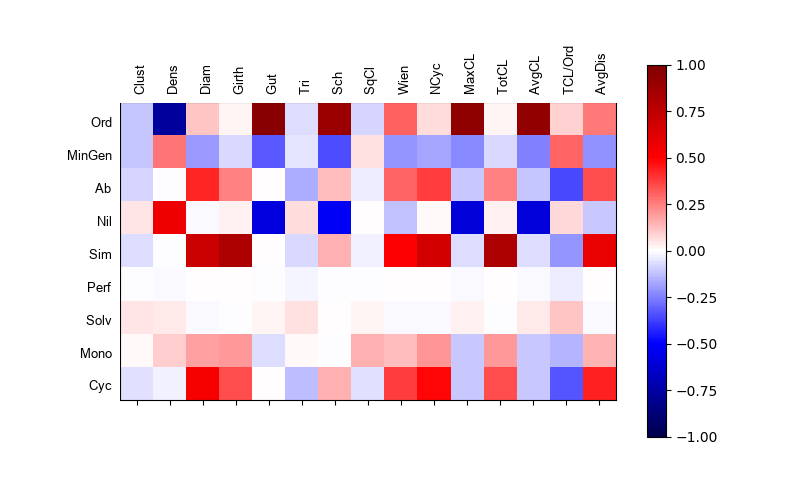}
    \caption{Top left: correlation matrix for pairs of graph statistics. Top right: correlation matrix for pairs of group properties. Bottom: correlation matrix between graph statistics and group properties. Abbreviations: Ord = group order, MinGen = minimum generators, Ab = abelian, Nil = nilpotent, Sim = simple, Perf = perfect, Solv = solvable, Mono = monolithic, Cyc = cyclic; Clust = average clustering coefficient, Dens = density, Diam = diameter, Girth = girth, Gut = Gutman index, Tri = triangles per node, Sch = Schultz index, SqCl = square clustering coefficient, Wien = Wiener index, NCyc = number of cycles, MaxCL = maximum cycle length, TotCL = total cycle lengths, AvgCL = average cycle length, TCL/Ord = total cycle lengths divided by group order, and AvgDis = average disorder.}
    \label{fig:overall_correlation}
\end{figure}

\begin{figure}
    \centering
    \includegraphics[width=0.7\linewidth]{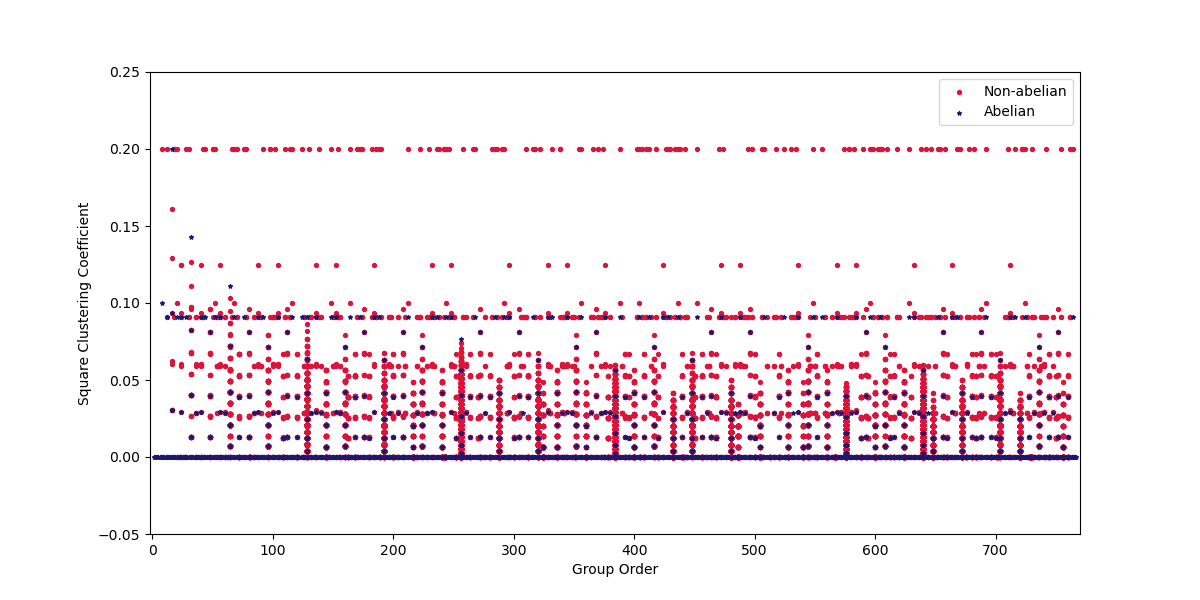}
    \caption{The square clustering coefficient for all groups in our dataset, coloured blue if the group is abelian and red otherwise.}
    \label{fig:abelian_square_cluster}
\end{figure}

\paragraph{Overall trends:}
We summarise Pearson correlation coefficients between all pairs of group and graph properties in Figure~\ref{fig:overall_correlation}, where for binary outcomes of group properties we denoted $1$ when a group possessed a given quality and 0 otherwise.
We were first surprised to observe no strong correlation between abelian groups and the square clustering coefficient (see Figure~\ref{fig:abelian_square_cluster}), since all generators should commute in these groups which would lead to a higher proportion of squares ($st=ts$ so that $g\xrightarrow{(g,gs)} gs\xrightarrow{(gs,gst)} g st=gts \xleftarrow{(gt,gts)} gt \xleftarrow{(g,gt)} g$ is a square).
We then realised that this was because we were using the \emph{directed} square clustering coefficient, which does not detect the (undirected) square corresponding to commuting pairs.
For this reason, we believe that the \textit{undirected} square clustering coefficient may be more expressive in detecting abelian groups (or weaker properties such as nilpotence).
However, we did observe that all perfect groups (nine total) had square clustering coefficient of zero, which means no collection of generators were able to form a directed square.
Indeed, we believe this is no coincidence, since perfect groups can be thought of as maximally non-abelian groups.
\begin{conjecture}
If $G$ is a perfect group, then $\square(\mathrm{Cay}(G,S))=0$ under appropriate restrictions to $S$.
\end{conjecture}

\begin{figure}
    \centering
    \includegraphics[width=0.4\linewidth]{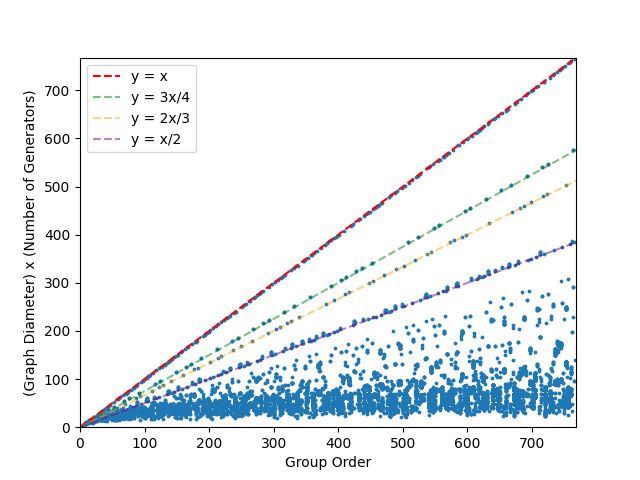}
    \includegraphics[width=0.4\linewidth]{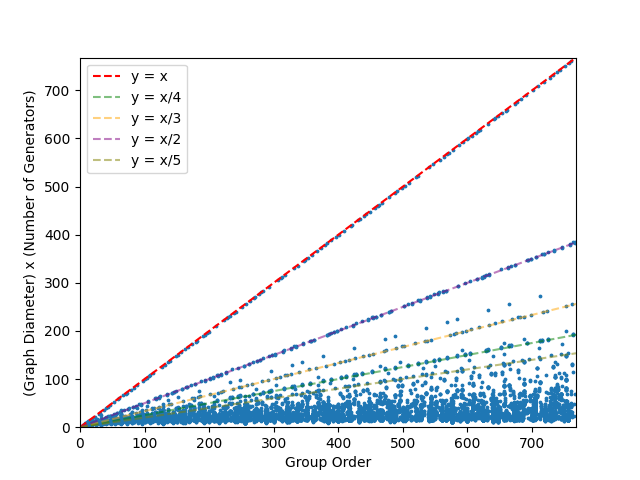}
    \caption{Left: Graph diameter multiplied by the number of generators plotted against group order for all our datapoints with various asymptotic trend lines highlighted. Right: The same plot, except with graph diameter multiplied by the \emph{minimum} number of generators for the group.}
    \label{fig:diam-gens}
\end{figure}

Next, we looked into the graph diameter, and found that this in general appeared to be bounded above by the ratio $n/g$, with the only exception for the three abelian groups of order $8$ ($(\Z/2\Z)^3$, $(\Z/2\Z) \times (\Z/4\Z)$ and $\Z/8\Z$) and equality achieved for a further 84 abelian groups (all the cyclic groups of order $2p$ for $p$ prime and all other abelian groups of order $4,12$ and $16$).
On further inspection, we also found an even stronger relation that the diameter is bounded above by $n/g_{\mathrm{min}}$ for $g_\mathrm{min}$ the minimal number of generators with the only exception for $(\Z/2\Z)^3$ and equality only achieved for $(\Z/2\Z)^2$ and $(\Z/2\Z)^4$.
Both relations are shown in Figure~\ref{fig:diam-gens}.
All exceptions have very similar form: a group of the form $(\Z/2\Z)\times G$ whose order is either small or has at most 2 prime factors -- where $\Z/2p\Z\simeq (\Z/2\Z)\times (\Z/p\Z)$ when $p\neq 2$.
This is likely related to the fact that by construction, these Cayley graphs of these are all product graphs of small cyclic graphs, and adding additional (non-abelian) group relations would only serve to reduce the number of independent directions the generators probe into the graph.
For example, the quaternion group $Q_8$ with three generators of order 2 has diameter 2 and not 3.
In Figure~\ref{fig:diam-gens} we also observe a number of different asymptotic trend lines for different families of groups, although we were unable to find any clear connections between groups on each trend line.
\begin{conjecture} \label{conj:gen-diam}
If $\mathfrak{D}(\mathrm{Cay}(G,S))\geq n/g_\mathrm{min}$ then $G=(\Z/2\Z)^k$ and $g=k$ for $k=2,3,4$.
Moreover, the value $\mathfrak{D}(\mathrm{Cay}(G,S)\cdot g_\mathrm{min}/n$ is directly related to some group property.
\end{conjecture}

\begin{figure}
    \centering
    \includegraphics[width=0.4\linewidth]{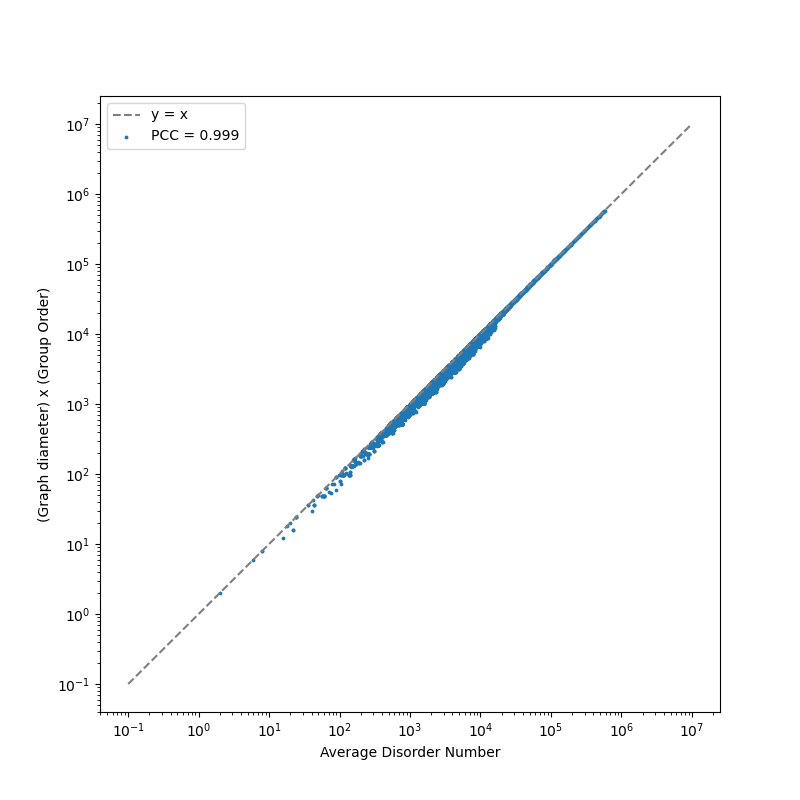}
    \caption{Group order multiplied by graph diameter plotted against the average graph disorder for all of our datapoints, which almost perfectly fits the line $y=x$.}
    \label{fig:ave_disorder_relationship}
\end{figure}

The most interesting correlation we observed was between average graph disorder and graph diameter.
On further inspection when plotting graph disorder against diameter, we observed several near perfect curves which appeared to depend on group order.
When multiplying graph diameter by group order, we then observed an almost perfect linear relationship with the average disorder, shown in Figure~\ref{fig:ave_disorder_relationship}.
The value $n\,\mathfrak{D}(\mathrm{Cay}(G,S))$ is always less than or equal to $\mathcal{A}(\mathrm{Cay}(G,S))$. To our knowledge, this is a novel discovery.
We also suspect the difference $n\,\mathfrak{D}(\mathrm{Cay}(G,S))-\mathcal{A}(\mathrm{Cay}(G,S))$ is related to generators of order two.
\begin{conjecture}
Under appropriate conditions, there is $\epsilon(\mathrm{Cay}(G,S)) = o(\mathcal{A}(\mathrm{Cay}(G,S)))$ which is a function of the number of generators in $S$ of order $2$, so that
\begin{equation}
n\,\mathfrak{D}(\mathrm{Cay}(G,S)) = \mathcal{A}(\mathrm{Cay}(G,S))-\epsilon(\mathrm{Cay}(G,S)).
\label{eq:weird_relation}
\end{equation}
\end{conjecture}

\begin{figure}
    \centering
    \includegraphics[width=0.6\linewidth]{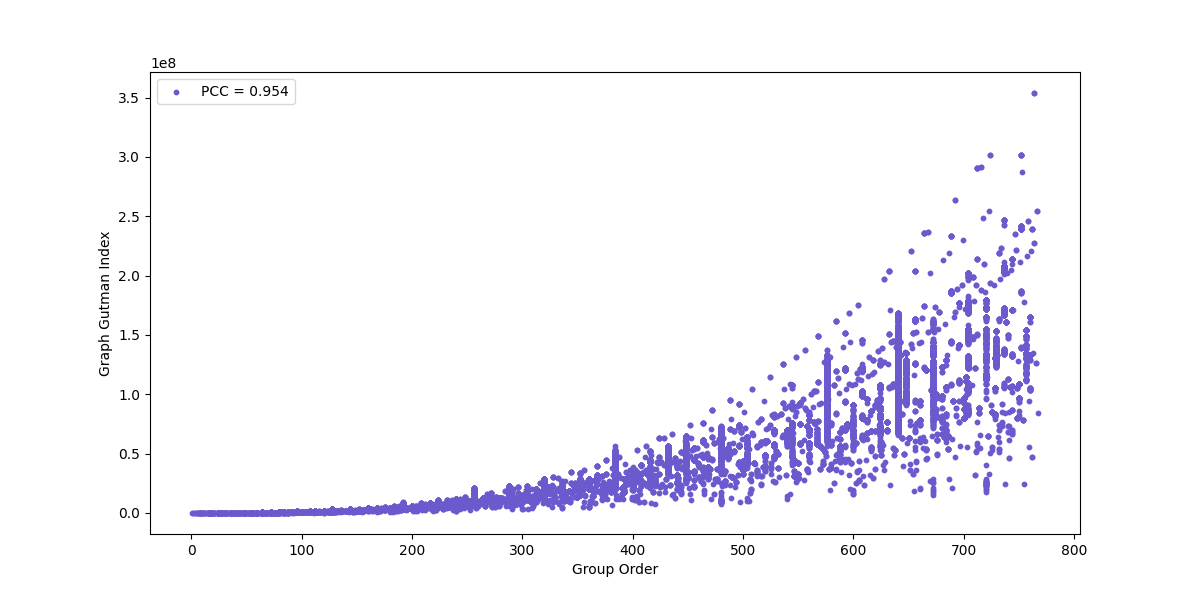}
    \includegraphics[width=0.3\linewidth]{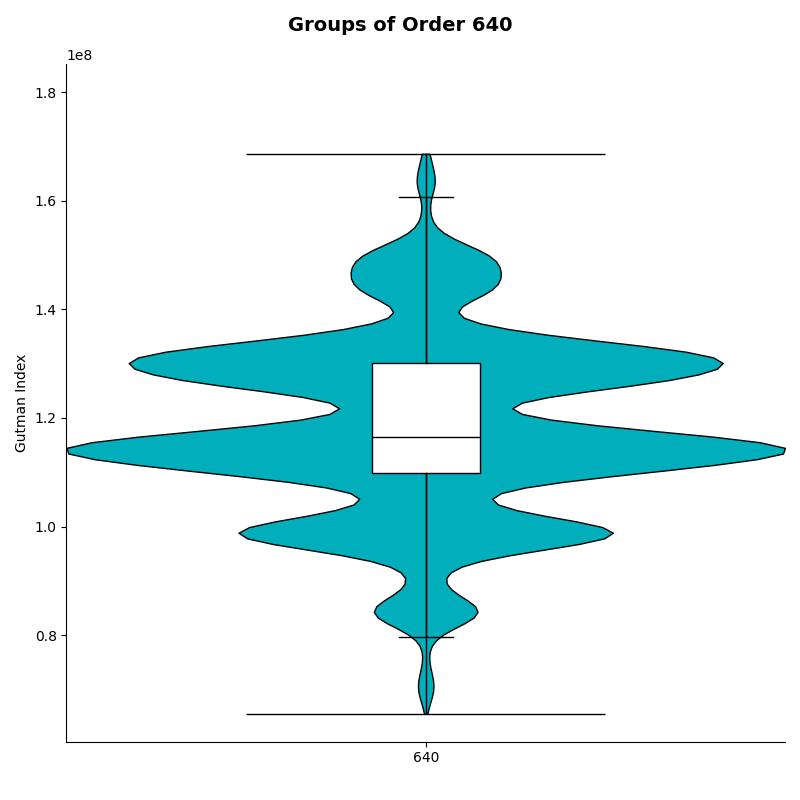}
    \caption{Left: plotting Gutman index against group order for all groups in our dataset. Right: violin plot demonstrating the multimodal Gutman index distribution for groups of order 640.}
    \label{fig:gutman-continuum}
\end{figure}

We also found that Gutman and Schultz indices (and the Wiener index to a lesser extent) are strongly correlated to the group order.
This is as expected (higher group order means more nodes means more pairs to sum over), but interestingly, we observed that for fixed group orders with a large numbers of groups there was a large spread in (see e.g. the vertical lines appearing throughout Figure~\ref{fig:gutman-continuum}).
On further inspection, we also found these were multimodally distributed.

Some of the observed (anti-)correlations can be explained by more subtle reasoning.
For instance, nilpotence was found to be anticorrelated with group order and number of generators, which we attribute to the shear number of groups whose order is not a prime power but has many prime factors (e.g. 17,841 out of 20,169 groups of order 384, 8,174 out of 8,681 groups of order 576, and 19,213 out of 21,541 groups of order 640 are non-nilpotent).
Similarly, correlations between girth, diameter and number of cycles are likely biased by large group order (and/or cyclic groups; see below).
We also see anticorrelation between Gutman and Schultz indices (and to a lesser extent the Wiener index) and graph density, which we can attribute to higher density graphs implying shorter paths between nodes since a greater proportion of edges exist.

\begin{figure}
    \centering
    \includegraphics[width=0.3\linewidth]{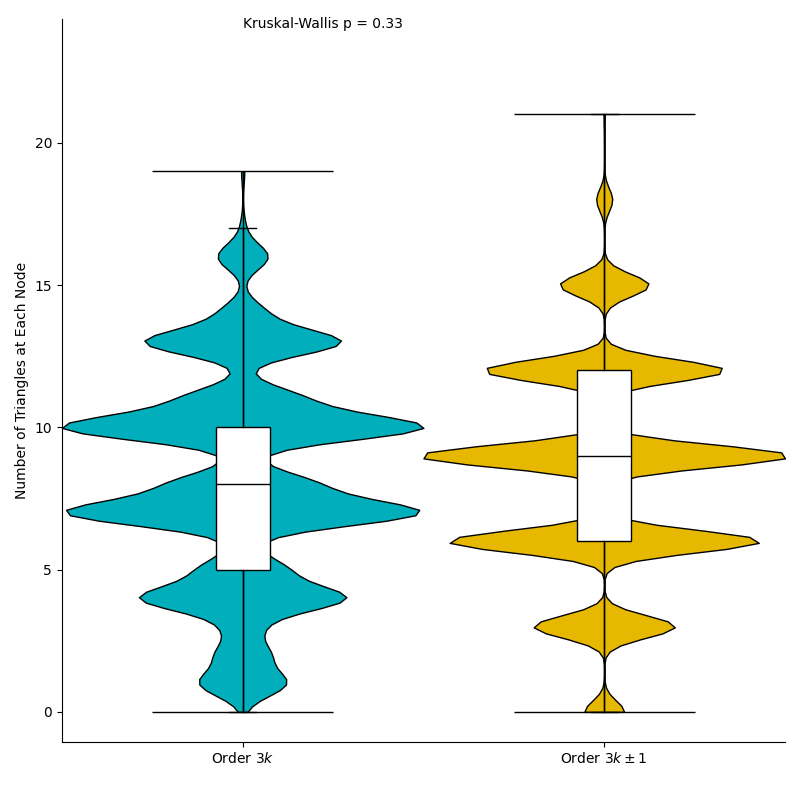}
    \caption{Violin plots comparing the number of triangles per node for groups whose order is a multiple of three (left) to those whose order is not a multiple of three (right).}
    \label{fig:order_3k}
\end{figure}

Finally, we were interested to see if group generators of order three biased any correlations as these would always contribute a (directed) triangle $\{(g,gs),(gs,gs^2),(gs^2,g)\}$ to every node.
Simple group theory implies this is only an issue with groups whose order is a multiple of three, so we compared the distribution of such groups to those whose order is not a multiple of three, which we show in Figure~\ref{fig:order_3k}.
There was little variation in the mean, standard deviation and range of both collections of groups (with no statistically significant difference from a Kruskal-Wallis test), although we found that groups whose order was not a multiple of three were \textit{only} able to have a multiple of three triangles at each node.
This was initially unexpected, but we later realised that this was because if there is a directed triangle in $\mathrm{Cay}(G,S)$ corresponding to the multiple $rst=e$ of generators $r,s,t$ (where not all three of $r,s,t$ are the same, and $e$ is the identity) then $\{(g,gr),(gr,grs),(grs,g)\}$, $\{(g,gs),(gs,gst),(gst,g)\}$ and $\{(g,gt),(gt,gtr),(gtr,g)\}$ would all create distinct directed triangles attributed to the node $g$ (the latter two corresponding to the relations $str=e$ and $trs=e$, respectively).

\begin{figure}
    \centering
    \includegraphics[width=0.4\linewidth]{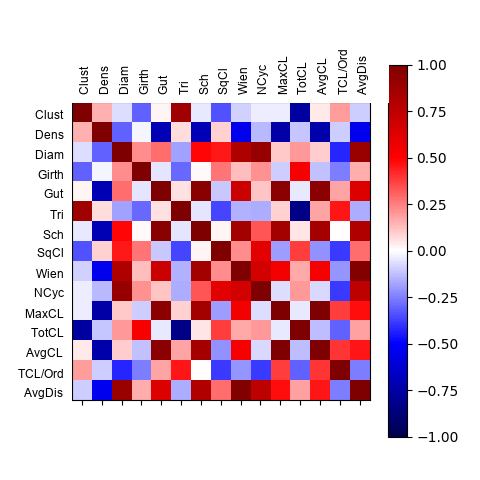}
    \includegraphics[width=0.4\linewidth]{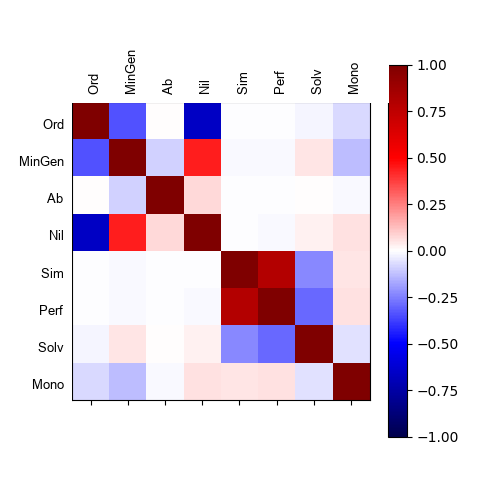}
    \includegraphics[width=0.6\linewidth]{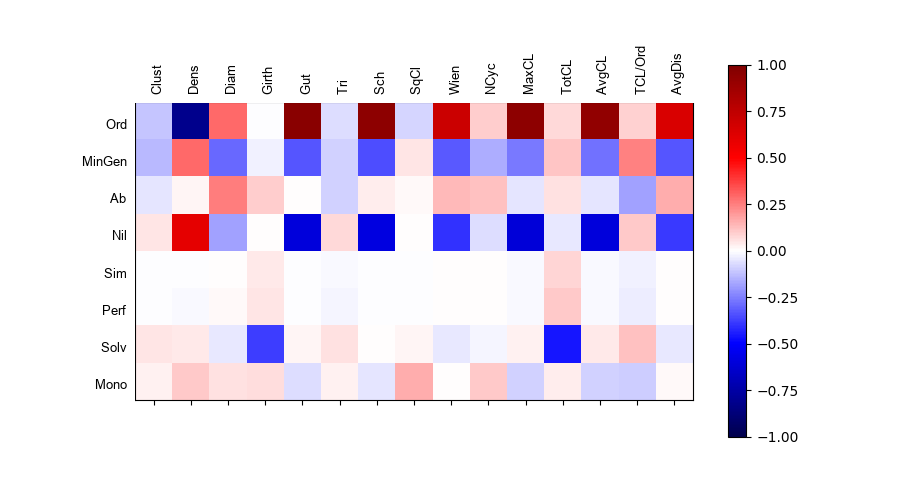}
    \caption{Correlation matrices as in Figure~\ref{fig:overall_correlation} restricted to non-cyclic groups in our dataset. Top left: for pairs of graph statistics. Top right: for pairs of group properties. Bottom: between pairs of graph statistics and group properties. Abbreviations: Ord = group order, MinGen = minimum generators, Ab = abelian, Nil = nilpotent, Sim = simple, Perf = perfect, Solv = solvable, Mono = monolithic; Clust = average clustering coefficient, Dens = density, Diam = diameter, Girth = girth, Gut = Gutman index, Tri = triangles per node, Sch = Schultz index, SqCl = square clustering coefficient, Wien = Wiener index, NCyc = number of cycles, MaxCL = maximum cycle length, TotCL = total cycle lengths, AvgCL = average cycle length, TCL/Ord = total cycle lengths divided by group order, and AvgDis = average disorder.}
    \label{fig:noncyclic_correlation}
\end{figure}

\paragraph{Removing cyclic groups:}
Several of the correlations we observed in the whole dataset (most notably for cyclic and simple groups) are biased by cyclic groups of prime order (and sometimes non-prime order) which are all represented by large cyclic graphs in our dataset whose properties are easily deduced.
For this reason, we also restricted our analysis to non-cyclic groups in our dataset.
Figure~\ref{fig:noncyclic_correlation} shows the correlation coefficients between these restricted datasets.

As before, we see a number of correlations explainable by group orders and/or graph combinatorics.
A clear example of this is for solvable groups, which have relatively strong anticorrelation between girth and the total summed length of the cycle basis, but is biased by the fact that there are so few non-solvable groups with small group order.
Indeed, after normalising the summed cycle lengths by the group order we instead observe a (faint) positive correlation.
Similarly, group order explains the observed anticorrelations between nilpotence and total cycle length, as well as the number of triangles per node and the summed cycle length, and further explains the other strong anticorrelations between nilpotence and graph statistics.

\begin{figure}
    \centering
    \includegraphics[width=0.4\linewidth]{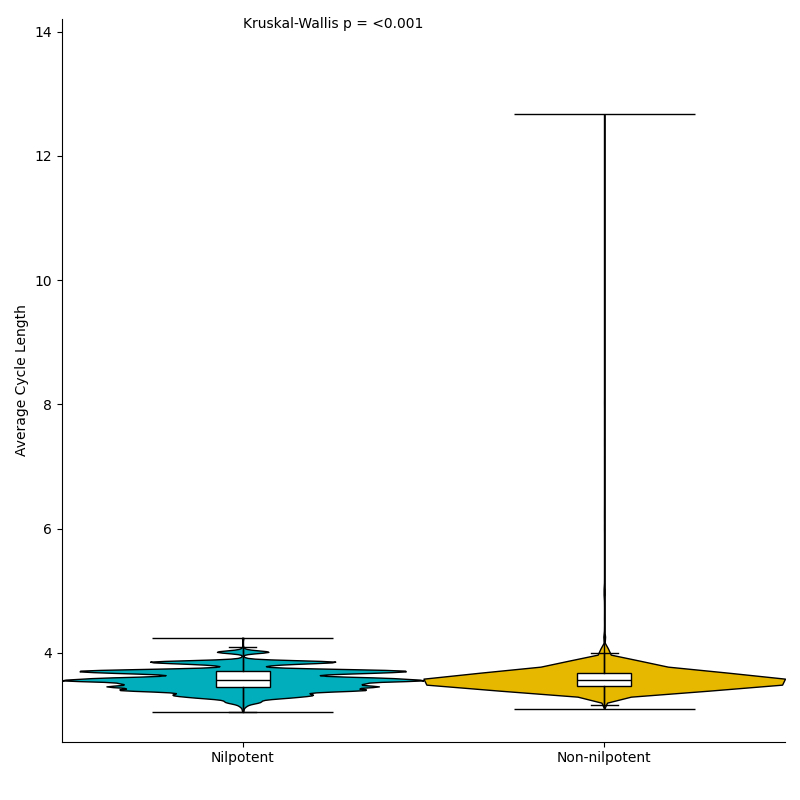}
    \includegraphics[width=0.4\linewidth]{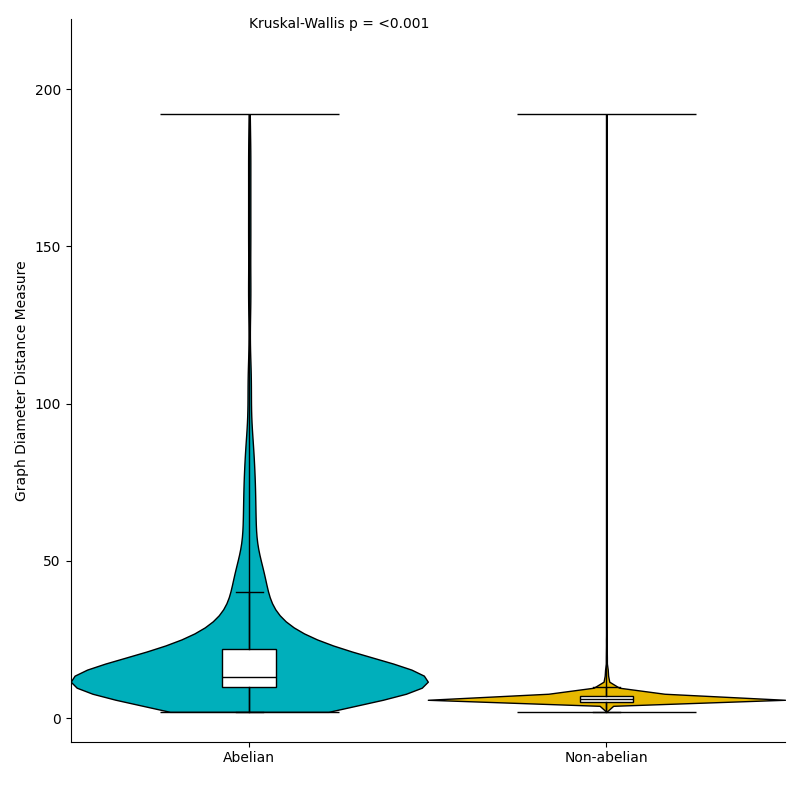}
    \caption{Left: violin plot comparing the distribution of the average length of cycles in the minimal cycle basis for non-cyclic nilpotent and non-nilpotent groups in our dataset. Right: violin plot comparing the distribution of graph diameter for non-cyclic abelian and non-abelian groups.}
    \label{fig:non-cyclic-plots}
\end{figure}

Looking into nilpotent groups more deeply, we found that while the mean and variations of average cycle basis length for nilpotent and non-nilpotent graphs is very similar, the distribution for nilpotent graphs has several quantised peaks (see Figure~\ref{fig:non-cyclic-plots}) which do not appear to correspond to large numbers of groups of fixed order.
We also still observe a mild correlation of non-cyclic abelian groups with graph diameter.
The distributions for the diameter of non-cyclic abelian and non-abelian groups is in Figure~\ref{fig:non-cyclic-plots}, where we see that the majority of abelian groups have diameter larger than the peak in the distribution for non-abelian groups.
We believe that this difference may relate to Conjecture~\ref{conj:gen-diam} and be explained by abelian groups with few generators whose Cayley graphs are product graphs, and this is only more prominent than for non-abelian groups because there are so few in comparison.

Finally, we observed two trends about the square clustering coefficient in non-cyclic groups.
First, there is a mild correlation between the square clustering coefficient and the graph diameter which we believe may also be related to Conjecture~\ref{conj:gen-diam}, or more generally for groups which factor as $G(\Z/2\Z)$.
Second, we noticed that all non-cyclic simple groups have zero square clustering coefficient, but this is of course because they are known to be perfect groups.

%\begin{itemize}
    %\item Some trivial/explainable correlations
    %\begin{itemize}
        %\item Non-solvable groups are correlated with total cycle length (biased by group order)normalising by group order it is almost identical
    %\end{itemize}
    %\item Strong anticorrelation between triangles per node and total cycle length (more smaller cycles)
    %\item Non-solvable groups have larger disorder on average, probably biased by larger group sizes
    %\item Nilpotent has much smaller total cycle length than non-nilpotent but after 
    %\item Non-cyclic abelian groups still have mild correlation with graph diameter
    %\item Mild correlation between diameter and square clustering coefficient
    %\item Solvable/perfect/simple groups
    %\begin{itemize}
        %\item Non-solvable/Perfect/simple groups appear to have larger girth (and very few triangles if any) and larger diameter
        %\item Non-solvable/perfect/simple groups have fewer cycles which are longer on average, and lower total cycle length when normalised by group size 
    %\end{itemize}
    %\item Non-cyclic simple groups have square clustering coefficient zero - but not sure if this is affected by generators of order 2
    %\item Nilpotent
    %\item Similar average cycle length, but nilpotent average cycle length appears more quantised than non-nilpotent
    %\begin{itemize}
        %\item Nilpotent has much lower average disorder and Gutman on average
    %\end{itemize}
%\end{itemize}

\begin{figure}
    \centering
    \includegraphics[width=0.35\linewidth]{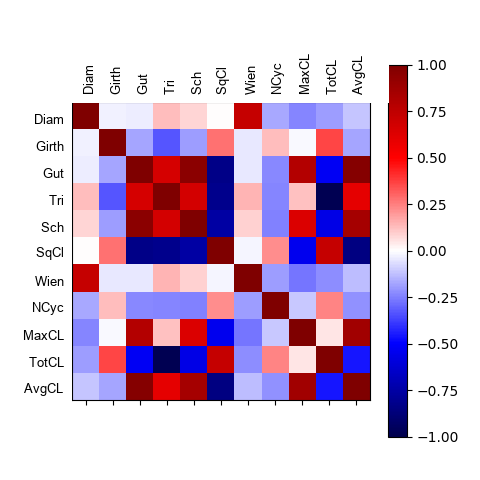}
    \includegraphics[width=0.35\linewidth]{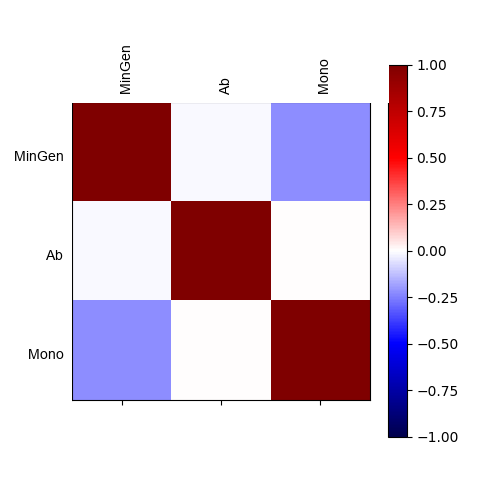}
    \includegraphics[width=0.6\linewidth]{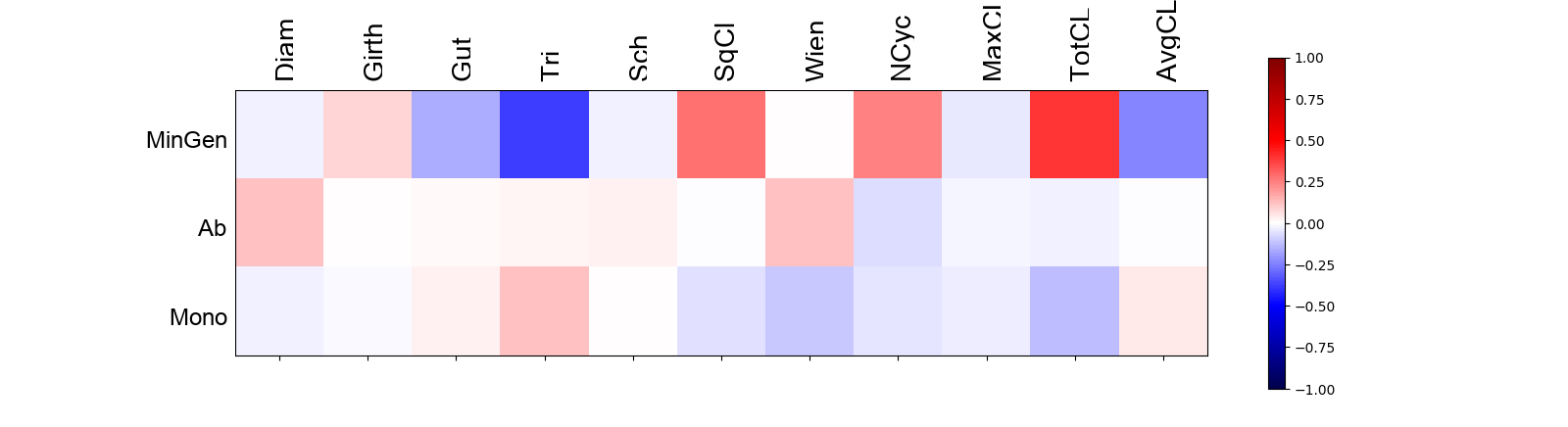}
    \caption{Correlation matrices as in Figure~\ref{fig:overall_correlation} for non-trivial properties of groups of order 256 in our dataset. Top left: for pairs of graph statistics. Top right: for pairs of group properties. Bottom: between pairs of graph statistics and group properties. Abbreviations: MinGen = minimum generators, Ab = abelian, Mono = monolithic; Diam = diameter, Girth = girth, Gut = Gutman index, Tri = triangles per node, Sch = Schultz index, SqCl = square clustering coefficient, Wien = Wiener index, NCyc = number of cycles, MaxCL = maximum cycle length, TotCL = total cycle lengths, and AvgCL = average cycle length.}
    \label{fig:256_correlation}
\end{figure}

\paragraph{Order 256:}\label{sec:order-256-analysis}
Finally, as groups of order 256 make up such a high proportion of our dataset, we also analysed correlations between only groups of order 256.
There is only one simple (and cyclic) group, all groups are nilpotent (and therefore solvable), and there are no perfect groups, so we only focus on abelian (22 of 56,092 groups) and monolithic groups (1,130 of 56,092 groups).
Similarly, many network statistics (depending on group order and/or number of generators) are redundant due to how these groups are generated.
We summarise the remaining correlation coefficients in Figure~\ref{fig:256_correlation}.

All groups of order 256 in our dataset have 8 generators, so we first looked at the trends related to the minimal number of generators.
We first notice a slight anticorrelation between the minimal number of generators of a group and whether it is monolithic, and on further inspection we found that the vast majority of monolithic groups can be generated by at most four elements (1,091 of 1,130 total) whereas the vast majority of non-monolithic groups must be generated by \emph{at least} four elements (48,968 of 54,962 total).
This is likely biased minimal number of generators of the centre; the smallest non-trivial normal subgroup of a monolithic group of order 256 must be contained in its centre, since all $p$-groups have non-trivial centre, which is itself an abelian normal subgroup. The centre therefore has order $2^k$ for $1\leq k\leq 8$, and the only abelian groups of this form which are monolthic are cyclic -- 8 total, compared to 16, 16, 12, 7, 4, 2 and 1 generated by a minimum of 2, 3, 4, 5, 6, 7 and 8 elements respectively, likely skewing non-monolithic groups to have more generators.

We also notice mild correlations with the summed length of cycles in the minimal cycle basis and an anticorrelation to the number of triangles.
The latter makes sense; more triangles means more group relations, which increases the chance of redundancy in the generating set.
In turn, the correlation to the summed cycle length is mediated by the strong anticorrelation between the number of triangles and the summed cycle length, which arises because we expect a fixed number of edges, so more triangles means fewer cycles of larger length that contribute to the total sum.

\begin{figure}
    \centering
    \includegraphics[width=0.49\linewidth]{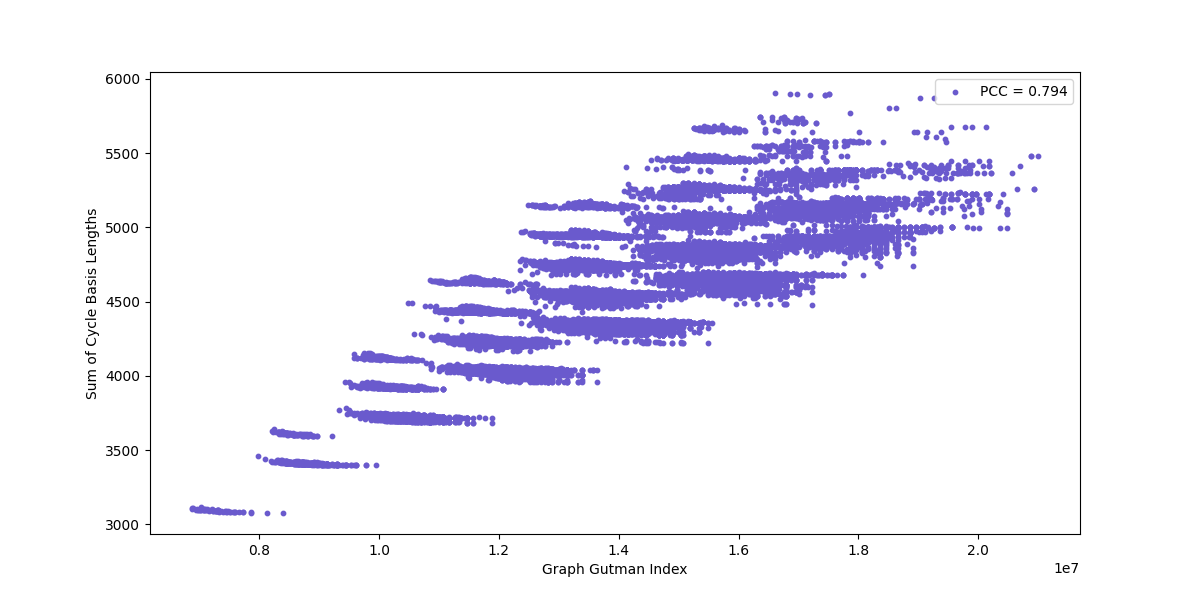}
    \includegraphics[width=0.49\linewidth]{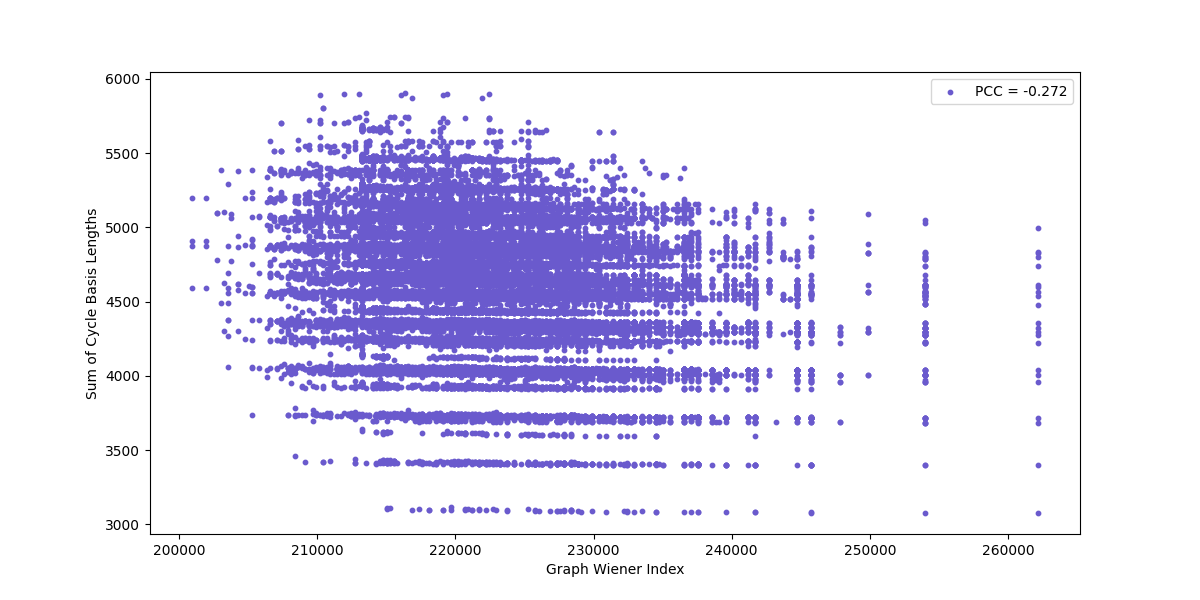}
    \caption{Correlation plots with Pearson correlation coefficient between the Gutman index and summed cycle lengths (left) and the Wiener index and summed cycle lengths (right) for groups of order 256.}
    \label{fig:256-index-correlations}
\end{figure}

The anticorrelation between summed cycle lengths and number of triangles also extends to explain a web of relationships between the square clustering coefficient, number of triangles per node and other cycle length statistics from the minimal cycle basis.
In addition, there are strong anticorrelations between the Gutman and Schultz indices and the summed cycle lengths, which are all relatively uncorrelated to the Wiener index, as shown in Figure~\ref{fig:256-index-correlations}.
We believe this is related to the number of generators of order two in our dataset.
On one hand, recall that the Wiener and Schultz index are proportional to $(2g-g_2)^2$ and $4g-2g_2$ respectively, and so are expected to be smaller as $g_2$ increases since $g=8$ is fixed in our dataset.
On the other hand, the more generators of order two there are, the fewer edges in the underlying undirected network and the more smaller cycles there are (e.g. every pair of generators of order two contributes two squares from their commutator relations), which decreases the number of cycles and the length of those cycles.

In terms of group properties, we see very few correlations between abelian and/or monolithic groups of order 256 and any network statistics.
The mild correlation between abelian groups and graph diameter is perhaps noteworthy given how few abelian groups there are, although this is again likely related to Conjecture~\ref{conj:gen-diam}.

\begin{figure}
    \centering
    \includegraphics[width=0.3\linewidth]{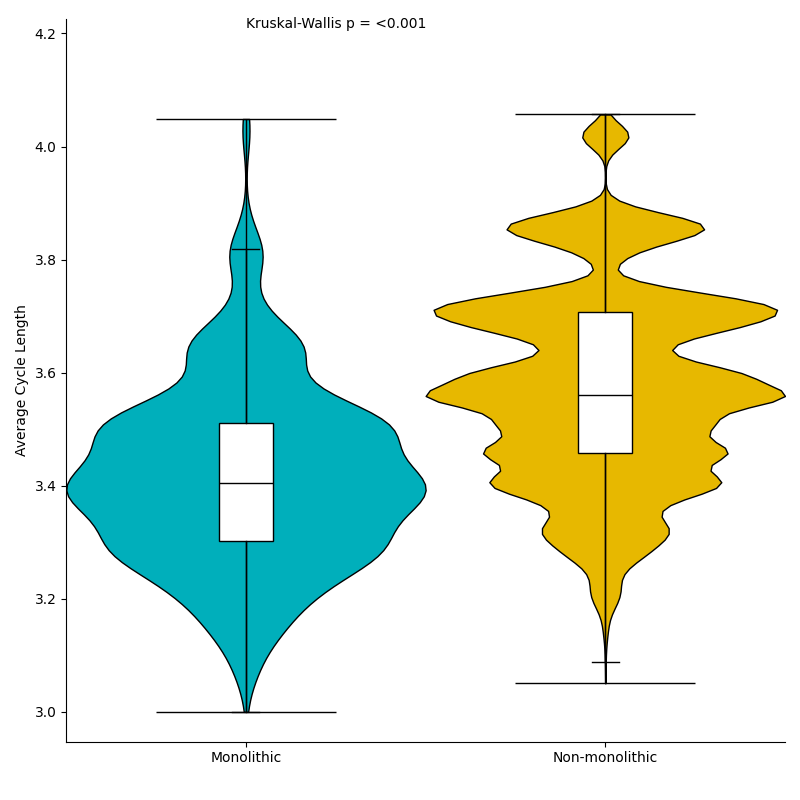}
    \includegraphics[width=0.3\linewidth]{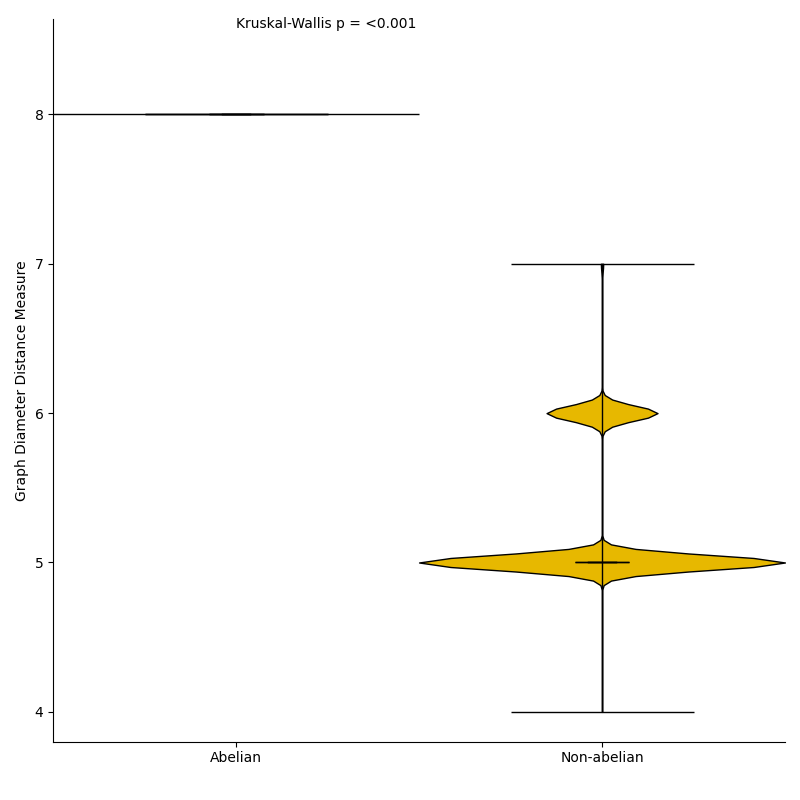}
    \caption{Left: violin plots comparing the distribution of average cycle lengths for monolithic and non-monolithic groups of order 256 in our dataset. Right: violin plots comparing the distribution of diameters for abelian and non-abelian groups of order 256 in our dataset.}
    \label{fig:256-group-distributions}
\end{figure}

Deeper analysis comparing distributions of graph statistics with and without each group property found that monolithic groups have smaller average cycle length on average than non-monolithic groups (see Figure~\ref{fig:256-group-distributions}).
However, the multimodal peaks in the distribution for non-monolithic groups suggests this is possibly related to having more group generators of order two (in particular, when the centre is non-cyclic and a product of copies of $\Z/2\Z$).
On the other hand, it was interesting to find all abelian groups of order 256 had Cayley graph diameter of 8, whereas all non-abelian groups had diameter at most 7 (see Figure~\ref{fig:256-group-distributions}).
In addition, abelian groups had a constant Wiener index of 262,144 which was greater than that for all non-abelian groups, where the maximum was 253,952.
While this gives us a complete discriminant for identifying abelian groups of order 256 in our dataset, it is not an inherent property of the groups in general. 
For example, $\mathrm{Cay}(\Z/256\Z,\{1\})$ has diameter 255 (with directed edges), and for the non-abelian group $G=Q_8\times \Z/ 32\Z$ (where $Q_8$ is the quaternion group of order 8 with generators $i,j,k$ and identity $e$) with generating set $S=\{(i,0),(j,0),(k,0),(e,1)\}$, $\mathrm{Cay}(G,S)$ has diameter 33.

\subsection{Spectral Analysis}
\label{sec:spectral}

The spectrum of the graph Laplacian provides a compact and informative summary of a network's structural and dynamical properties. For an undirected graph $H = (V, E)$ with $|V| = n$ vertices, adjacency matrix $A$, and diagonal degree matrix $D$, the combinatorial Laplacian is defined as $\mathbf{L} = \mathbf{D} - \mathbf{A}$. The matrix $\mathbf{L}$ is symmetric and positive semidefinite, so its eigenvalues are real and non-negative and can be ordered as $0 = \lambda_1 \leq \lambda_2 \leq \cdots \leq \lambda_n$. The multiplicity of the zero eigenvalue equals the number of connected components of $H$, while the remaining eigenvalues encode information about connectivity, expansion, diameter, and the dynamics of diffusion and synchronization processes supported on the graph~\cite{mohar1991laplacian,chung1997spectral,merris1994laplacian}. For $d$-regular graphs --- including the underlying undirected graph of all Cayley graphs, which are vertex-transitive and therefore regular --- the Laplacian admits a particularly clean form, $\mathbf{L} = d\mathbf{I} - \mathbf{A}$, so that the Laplacian and adjacency spectra are linearly related. This makes the rich machinery developed for adjacency spectra of regular graphs, including Cheeger-type inequalities and bounds for expander graphs, directly applicable~\cite{brouwer2012spectra,hoory2006expander}.

Because the present work compares Cayley graphs of all group orders in the range $1$ to $767$ and, in general, different degrees, we work throughout with the \emph{normalized Laplacian}
\begin{equation}
    \mathbf{L}_N = \mathbf{D}^{-1/2} \mathbf{L} \, \mathbf{D}^{-1/2} = \mathbf{I} - \mathbf{D}^{-1/2} \mathbf{A} \, \mathbf{D}^{-1/2},
\end{equation}
introduced by Chung~\cite{chung1997spectral}. The normalized Laplacian has eigenvalues confined to the interval $[0, 2]$ regardless of graph size or degree, which enables a well-defined, scale-invariant comparison of spectra across networks of heterogeneous size~\cite{chung1997spectral,banerjee2008spectrum,delange2014laplacian}. For a $d$-regular graph, the normalized and combinatorial Laplacians are simply related by $\mathbf{L}_N = \mathbf{L}/d$, so the normalization corresponds, in our setting, to expressing all spectral quantities in units of the common vertex degree. In what follows, $0 = \lambda_1 \leq \lambda_2 \leq \cdots \leq \lambda_n \leq 2$ denote the eigenvalues of $\mathbf{L}_N$. From each spectrum we extract five scalar descriptors: the Fiedler eigenvalue $\lambda_2$, the largest eigenvalue $\lambda_{\max}$, the expander measure $\lambda_2 / \lambda_{\max}$, the eigenvalue spread $\lambda_{\max} - \lambda_2$, and two complementary \emph{gap-index} statistics that locate the position of the most prominent eigengap in the spectrum. The latter are introduced in the spirit of the eigengap heuristic of spectral clustering~\cite{vonluxburg2007tutorial}, which interprets a pronounced gap between consecutive Laplacian eigenvalues as evidence of latent block structure in the graph.

\begin{figure}[t]
    \centering
    \includegraphics[width=\linewidth]{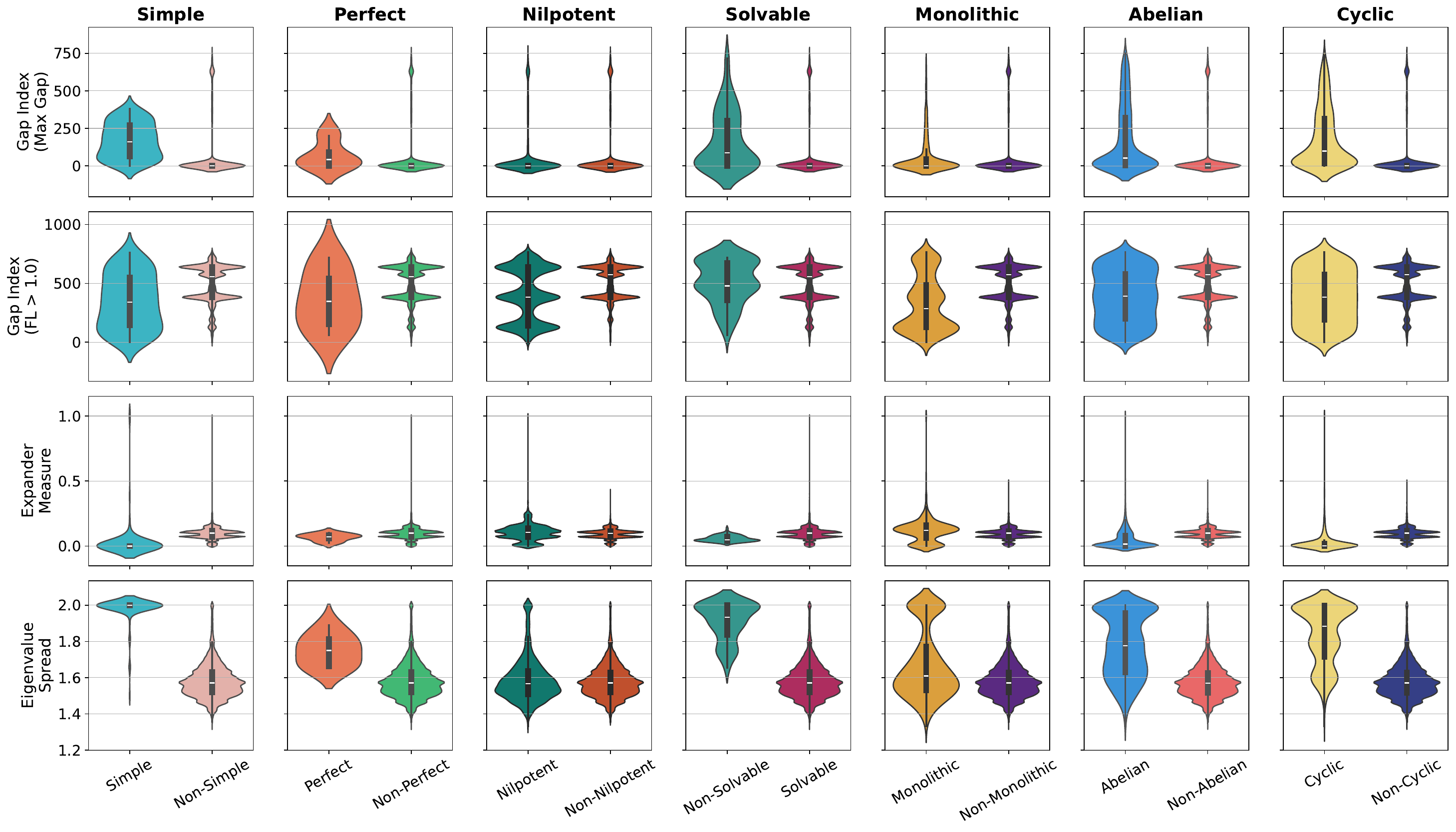}
    \caption{Distribution of four spectral descriptors of the normalised Laplacian of Cayley graphs, stratified by seven binary algebraic properties of the underlying group. The figure is organised as a $4 \times 7$ grid of paired violin plots. Each column corresponds to one algebraic property (\emph{simple}, \emph{perfect}, \emph{nilpotent}, \emph{solvable}, \emph{monolithic}, \emph{abelian}, \emph{cyclic}), and each row to one spectral descriptor: the max-gap index $k_{\max}$, the first-large-gap index $k_\tau$ at threshold $\tau = 1.0$, the expander measure $\lambda_2 / \lambda_{\max}$, and the eigenvalue spread $\lambda_{\max} - \lambda_2$. Within every cell, the left violin shows the distribution restricted to Cayley graphs whose underlying group satisfies the property, and the right violin shows the complementary class; the embedded box reports the median and interquartile range.}
    \label{fig:violin-by-property}
\end{figure}

\paragraph{Fiedler eigenvalue.}
The second-smallest eigenvalue $\lambda_2$ of the normalized Laplacian is known as the \emph{algebraic connectivity} or \emph{Fiedler eigenvalue}, after the seminal contribution of Fiedler~\cite{fiedler1973algebraic}. It satisfies $\lambda_2 > 0$ if and only if $H$ is connected, and its magnitude quantifies how strongly connected the graph is: graphs that are close to being disconnected, or that admit a sparse cut separating them into well-defined communities, exhibit a small Fiedler eigenvalue, whereas tightly knit graphs display large values of $\lambda_2$. This intuition is made precise by the Cheeger inequality, $\lambda_2 / 2 \leq \phi(H) \leq \sqrt{2 \lambda_2}$, where $\phi(H)$ is the edge conductance of $H$~\cite{alon1985isoperimetric,alon1986eigenvalues,chung1997spectral}. More specifically, $d$-regular graphs with $n$ nodes satisfy the inequality $\lambda_2\geq1-\sqrt{(n-d)/(nd-d)}$ (see for example \cite[Theorem~6]{abd2025eigenvalue}). The Fiedler eigenvalue also governs the rate of convergence of diffusion processes and consensus dynamics on the graph, and the corresponding eigenvector --- the Fiedler vector --- is the basis of standard spectral partitioning and clustering procedures~\cite{fiedler1975property,shi2000normalized,vonluxburg2007tutorial}.

\paragraph{Largest eigenvalue.}
The largest eigenvalue $\lambda_n$, also referred to as the \emph{Laplacian spectral radius}, lies in $[1, 2]$ for the normalized Laplacian of any connected graph and attains the upper bound $\lambda_n = 2$ if and only if $H$ has a bipartite connected component~\cite{chung1997spectral}. It captures the highest-frequency mode supported on the graph, and physically corresponds to the most oscillatory pattern of diffusion or vibration. From a dynamical standpoint, $\lambda_n$ controls the stability and admissible step-size of explicit Laplacian-based integrators, and the proximity of $\lambda_n$ to $2$ is a standard indicator of bipartiteness or near-bipartiteness of the underlying structure~\cite{chung1997spectral,mohar1991laplacian}. For Cayley graphs, $\lambda_n$ is particularly informative because the algebraic structure of the generating set often determines whether the graph is bipartite, and consequently whether $\lambda_n$ saturates its upper bound.

\begin{figure}[t]
    \centering
    \includegraphics[width=\linewidth]{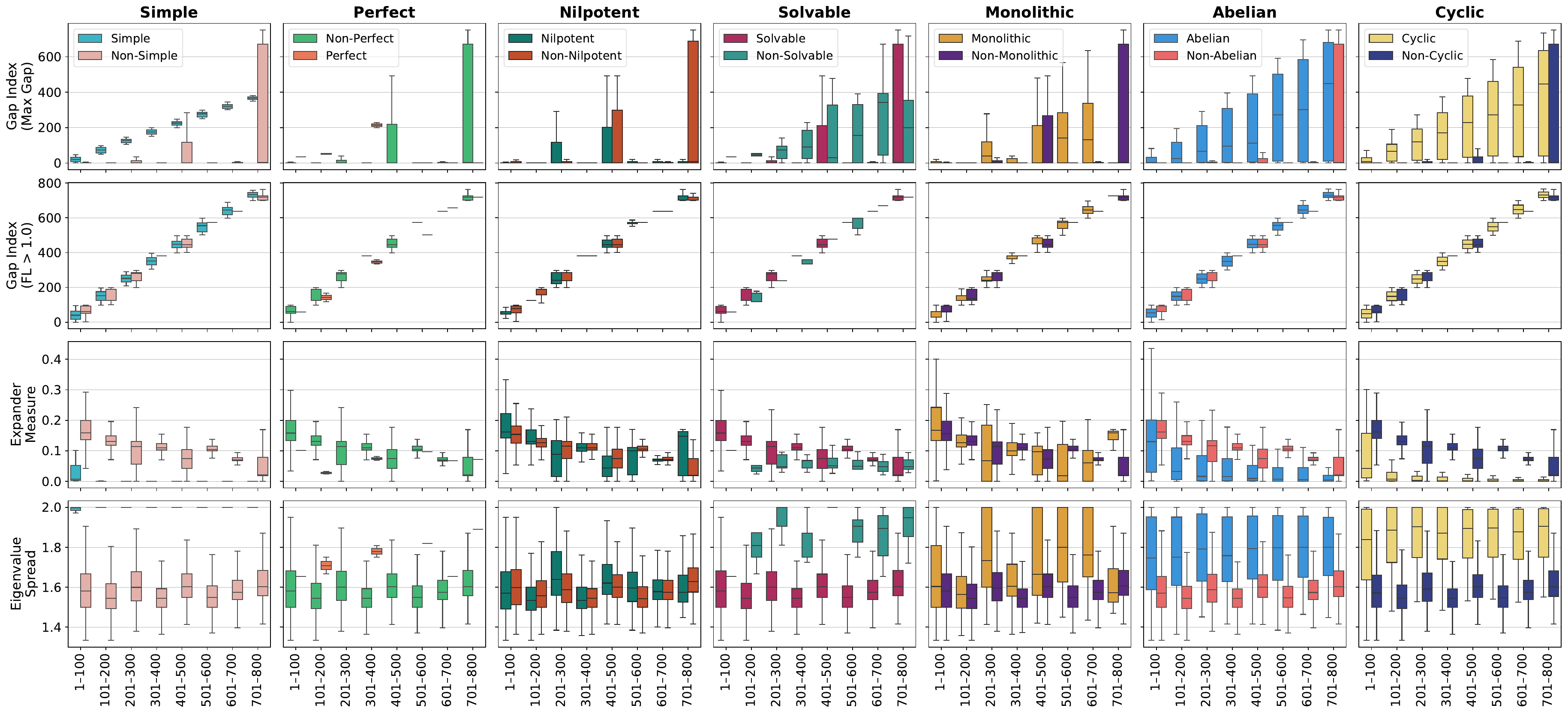}
    \caption{Dependence of the four spectral descriptors of Figure~\ref{fig:violin-by-property} on the order of the underlying group. Within each panel, groups are partitioned into bins of width $100$ (orders $1$--$100$, $101$--$200$, and so on), and the distribution of the descriptor in each bin is summarised by side-by-side box plots (outliers suppressed) for graphs satisfying and not satisfying the property.}
    \label{fig:boxplot-by-order}
\end{figure}

\paragraph{Expander measure.}
The dimensionless ratio $\lambda_2 / \lambda_n$ provides a scale-invariant indicator of how close a graph is to being an ideal expander. Values close to one are achieved by graphs whose spectrum is concentrated around a single bulk value, as is the case for complete graphs and, more generally, for Ramanujan-type expanders, while values close to zero indicate the presence of slow modes that hinder diffusion and signal the existence of bottlenecks or near-disconnections~\cite{hoory2006expander,chung1997spectral}. In the dynamical-systems literature, the reciprocal quantity $\lambda_n / \lambda_2$ --- sometimes called the \emph{Laplacian spectral ratio} --- was shown by Barahona and Pecora~\cite{barahona2002synchronization} to control the synchronizability of networks of identical diffusively coupled oscillators: a smaller $\lambda_n / \lambda_2$, equivalently a larger $\lambda_2 / \lambda_n$, corresponds to a wider stability window of the synchronized state. The expander measure thus simultaneously captures structural expansion and dynamical robustness, and is particularly well suited to comparisons across Cayley graphs of different sizes and degrees~\cite{you2012laplacian,lin2023laplacian}.

\paragraph{Eigenvalue spread.}
The \emph{Laplacian spread} is defined as $s(H) = \lambda_n - \lambda_2$, i.e.\ the width of the non-trivial portion of the spectrum~\cite{goldberg2006bounding,fan2008laplacian}. For the normalized Laplacian the spread takes values in $[0, 2]$ and admits an immediate interpretation as the dynamic range of timescales accessible to processes on the graph: a small spread indicates that all non-trivial modes evolve on comparable timescales, as is the case for highly homogeneous or expander-like graphs, whereas a large spread reveals a strong separation between the slowest and the fastest modes and is typically associated with structural heterogeneity, the presence of bottlenecks, or near-bipartite components. The Laplacian spread is therefore complementary to the expander measure: while $\lambda_2 / \lambda_n$ summarizes the relative balance between extremal eigenvalues, $s(H)$ quantifies their absolute separation on the normalized scale, making it informative even when both extrema move jointly across graphs of different sizes.

\paragraph{Gap-index statistics.}
In order to complement the extremal-eigenvalue descriptors introduced above with information about the \emph{internal} organisation of the spectrum, we further consider two indices that locate the position of the most prominent discontinuity in the ordered list of eigenvalues. Both descriptors are motivated by the eigengap heuristic~\cite{vonluxburg2007tutorial}: given the sorted spectrum of $\mathbf{L}_N$, define the consecutive eigengaps $\delta_i = \lambda_{i+1} - \lambda_i$ for $i = 1, \dots, n-1$. The first descriptor, which we refer to as the \emph{max-gap index}, is the index of the eigenvalue immediately preceding the largest eigengap, $k_{\max} = \arg\max_{i} \delta_i$, and corresponds to the standard data-driven choice for the number of latent clusters in spectral clustering. The second descriptor, which we refer to as the \emph{first-large-gap index}, is the smallest index at which the consecutive gap exceeds a fixed threshold $\tau$, $k_\tau = \min \{ i : \delta_i > \tau \}$, with $\tau = 1.0$ adopted throughout this work. Unlike $k_{\max}$, which is intrinsically robust because it depends only on the relative ordering of the gaps, $k_\tau$ is sensitive to the absolute scale of the spectrum and provides a complementary view that emphasises early, strong discontinuities. Comparing the two indices on the same spectrum is therefore a useful diagnostic of how concentrated the spectral mass is in the lower portion of the spectrum: the two indices coincide for graphs whose dominant gap is both early and large, but diverge when the spectrum exhibits multiple gaps of comparable magnitude.
\begin{figure}[t]
    \centering
    \includegraphics[width=\linewidth]{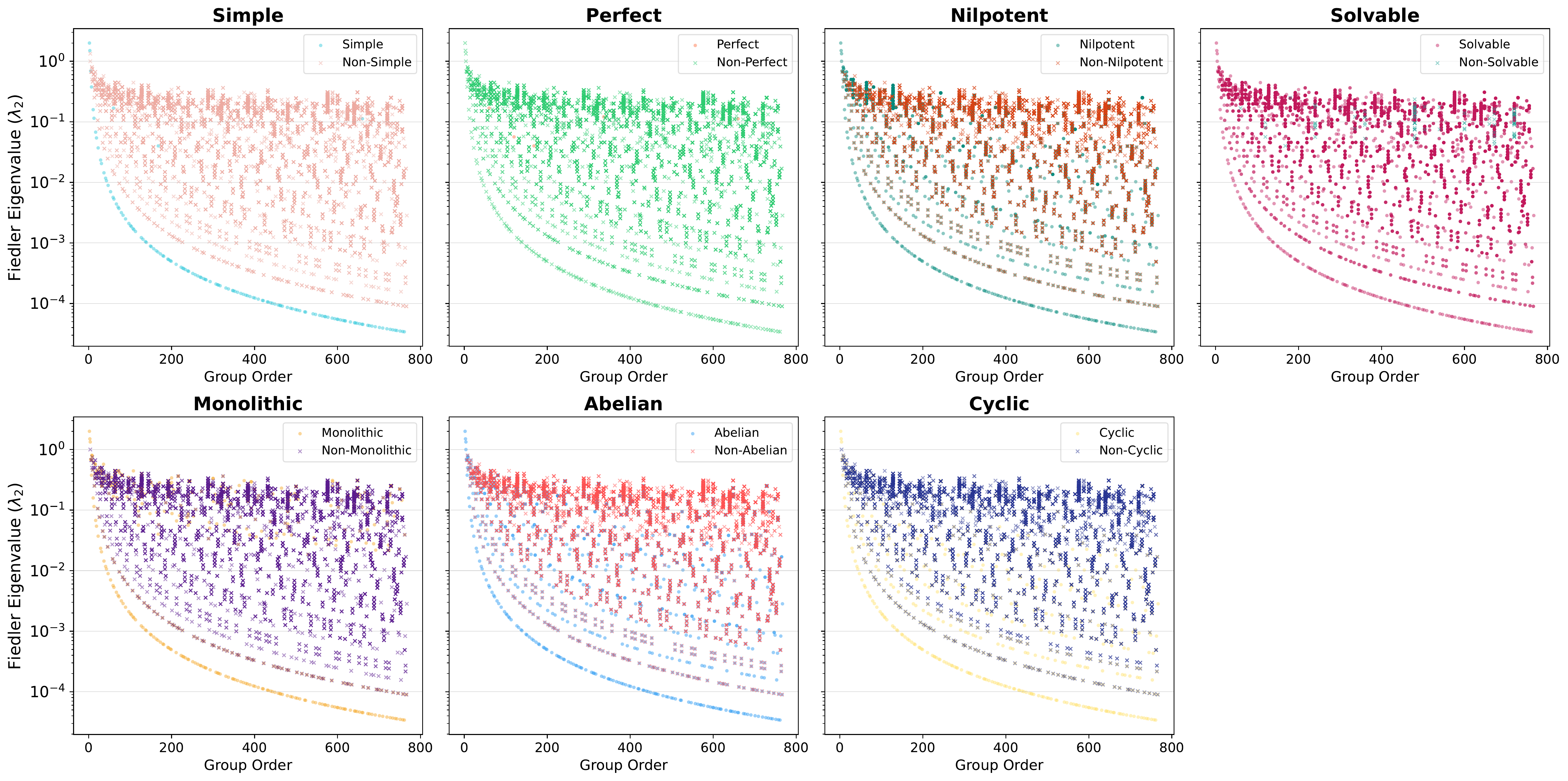}
    \caption{Fiedler eigenvalue $\lambda_2$ of the normalised Laplacian of each Cayley graph (in log scale) plotted against the exact order of the underlying group. The figure highlights how algebraic connectivity scales with group order and whether membership in a given property class is associated with systematically higher or lower connectivity.}
    \label{fig:fiedler-by-order}
\end{figure}

\paragraph{Empirical findings on Cayley graphs of small order.}
The five spectral descriptors introduced above are computed on the Cayley graphs of all finite groups of order between $1$ and approximately $767$, and their distributions are analysed against seven binary algebraic properties of the underlying group: \emph{simple}, \emph{perfect}, \emph{nilpotent}, \emph{solvable}, \emph{monolithic}, \emph{abelian}, and \emph{cyclic}. Figure~\ref{fig:violin-by-property} reports the per-property distribution of each measure as paired violin plots, Figure~\ref{fig:boxplot-by-order} stratifies the same comparison by group-order bins of width $100$, and Figure~\ref{fig:fiedler-by-order} shows the Fiedler eigenvalue against the exact group order on a logarithmic vertical axis. A first observation, visible across all three figures, is that the nilpotent and non-nilpotent classes produce remarkably similar distributions for four of the five measures; the only descriptor on which a clear separation emerges is the first-large-gap index $k_\tau$, whose distribution becomes markedly trimodal once restricted to the nilpotent subclass. This indicates that nilpotency is largely invisible to extremal-eigenvalue statistics but is detected by the internal structure of the spectrum, where it manifests as a quantised set of admissible positions for the first dominant eigengap. The trimodality is consistent with the layered subgroup structure of nilpotent groups, whose Cayley graphs inherit a multi-scale block organisation that is naturally reflected in the location of the first large discontinuity.

\begin{conjecture}[Central-quotient eigengaps]
Let $G$ be a finite nilpotent group with upper central series
\[
1=Z_0(G)\leq Z_1(G)\leq \cdots \leq Z_c(G)=G,
\]
and let $\Gamma=\mathrm{Cay}(G,S)$ be the underlying undirected Cayley graph associated to a chosen generating set $S$. If $k_{>1}=\min\{i:\lambda_{i+1}-\lambda_i>1\}$ is defined for the normalised Laplacian spectrum of $\Gamma$, then either $k_{>1}=|G|-1$, or
\[
k_{>1}=|G/Z_j(G)|
\]
for some $1\leq j\leq c$.
\end{conjecture}

This is plausible because, for any normal subgroup $N\triangleleft G$, functions on $G$ that are constant on cosets of $N$ form a $|G/N|$-dimensional subspace invariant under the Cayley graph operators, so quotient-scale modes are natural candidates for producing visible breaks in the ordered spectrum.

A second, broader pattern concerns the symmetry between the seven binary partitions. The graphs falling into the classes \emph{non-simple}, \emph{non-perfect}, \emph{non-nilpotent}, \emph{solvable}, \emph{non-monolithic}, \emph{non-abelian}, and \emph{non-cyclic} exhibit very similar spectral signatures to one another, both in central tendency and in dispersion, whereas their respective counterparts --- \emph{simple}, \emph{perfect}, \emph{nilpotent}, \emph{non-solvable}, \emph{monolithic}, \emph{abelian}, \emph{cyclic} --- display more pronounced variability and more heterogeneous distributions. In other words, the ``generic'' side of each algebraic dichotomy occupies a comparatively narrow region of spectral-descriptor space, while the ``structured'' side produces a broader, multi-modal distribution. A noteworthy exception is provided by the abelian and cyclic classes, whose spectral fingerprints are nearly indistinguishable from one another, in agreement with the fact that all cyclic groups are abelian and that the additional commutativity constraints imposed in passing from the abelian to the cyclic class do not appreciably alter the gross spectral shape of the associated Cayley graphs.

A more quantitative summary is obtained by examining the descriptors individually. The eigenvalue spread is systematically lower for the ``generic'' side of each partition than for its structured counterpart, indicating that groups with rich algebraic structure tend to produce Cayley graphs whose normalised spectra span a wider portion of the available interval $[0, 2]$. The expander measure, in contrast, displays comparable median values across both sides of every partition, suggesting that the relative balance between $\lambda_2$ and $\lambda_n$ is, to a first approximation, a property of the Cayley-graph construction itself rather than of the algebraic class of the generating group. The two gap-index statistics behave differently from one another: the max-gap index $k_{\max}$ and the first-large-gap index $k_\tau$ yield very similar distributions on the generic side of each partition, but diverge substantially on the structured side. This indicates that, for structured groups, the choice of threshold $\tau$ has a non-negligible effect on the inferred number of latent spectral clusters --- a useful empirical caveat for any subsequent partitioning analysis built on these graphs. The finer-grained stratification by group order, reported in Figure~\ref{fig:boxplot-by-order}, confirms the same qualitative pattern at every scale considered, ruling out the possibility that the global patterns are dominated by a small number of large groups. Finally, Figure~\ref{fig:fiedler-by-order} shows that, although the Fiedler eigenvalue spans several orders of magnitude as the group order grows, its distribution within each algebraic partition remains homogeneous between the two classes: membership in a given property class is not systematically associated with higher or lower algebraic connectivity, but rather with a similar dependence on group order modulated by class-specific fluctuations.

%%%%%%%%%%%%%%%%%%%%%%%%%%%%%%%%%%%%%%%%%%%%%%%%%
\section{Machine Learning}
\label{sec:ML}

In this section we evaluate to what extent the algebraic properties recorded in Section~\ref{sec:dataset_gen} can be predicted from Cayley-graph data alone. Four families of models are compared on the seven binary classification targets: abelian, nilpotent, simple, perfect, solvable, monolithic, and cyclic. These models are a logistic-regression baseline, two ensemble tree models (Random Forest and XGBoost), a feed-forward multilayer perceptron (MLP) operating on the edge list, and two graph neural network architectures (GCN and GIN) operating on the Cayley graph as a relational object; hyperparameters for each model are given with the respective \href{https://github.com/Engrima18/CayleyNet}{\texttt{GitHub}}. The classical models and the MLP operate on fixed-dimensional vector inputs and serve as a graded sequence of baselines: the logistic regression and tree models quantify the algebraic signal already captured by hand-crafted network observables, while the MLP quantifies the signal recoverable from the raw adjacency information when no graph-structured inductive bias is imposed. The GNNs, in contrast, exploit the graph structure explicitly through message passing and probe whether topology-aware representations extract additional signal beyond what either summary statistics or the raw adjacency matrix afford. To the best of our knowledge, this is the first benchmark of message-passing GNNs trained directly on Cayley graphs as labelled mathematical data.

\subsection{Selected Models}

\paragraph{Classical baselines.}
The logistic regression, Random Forest, and XGBoost classifiers take as input the vector of network statistics introduced in Section~\ref{sec:net_stats} — density, diameter, girth, Wiener/Gutman/Schultz indices, average and square clustering coefficients, number of triangles per node, and the summary statistics of the minimal cycle basis (maximum, sum, and mean of basis cycle lengths) — together with the minimal number of generators of the group.

\paragraph{MLP baseline.}
To isolate the value of imposing a graph-structured inductive bias, we additionally train a simple feed-forward neural network on the edge list of each Cayley graph. Given $\mathrm{Cay}(G,S)$ of order $n=|G|$ and $m$ edges, the input to the MLP is the vector $\mathrm{vec}(\mathbf{A})\in\{0, \dots, n-1\}^{2m}$ which represents the edge list of the graph. Because the dataset spans graphs of variable size, all edge lists are zero-padded to the maximum number of edges in the dataset; this is a deliberately naïve representation that does not respect the permutation symmetry of nodes (the MLP sees a specific vertex ordering given by GAP) and does not exploit the locality of the graph topology. A small comparison against this baseline, therefore, measures the signal recoverable by an unstructured neural network from the same raw information consumed by the GNNs, and any gap in favour of the GNNs is attributable to the message-passing inductive bias rather than to deep-learning expressivity alone.

\paragraph{Graph neural networks.}
The Graph Convolutional Network (GCN) \cite{kipf2017semi} and Graph Isomorphism Network (GIN) \cite{xu2018powerful} architectures operate directly on the Cayley graph as a relational object, and admit two equivalent and complementary descriptions that we briefly recall, since both inform our analysis.

From the \emph{message-passing} perspective, each layer of a GNN updates node representations $\mathbf{h}^{(\ell+1)}_v$ from those of their neighbours by a permutation-invariant aggregation followed by a learnable transformation,
\begin{equation}
    \mathbf{h}^{(\ell+1)}_v \;=\; \phi^{(\ell)}\Bigl(\mathbf{h}^{(\ell)}_v,\ \bigoplus_{u\in\mathcal{N}(v)} \psi^{(\ell)}\bigl(\mathbf{h}^{(\ell)}_u\bigr)\Bigr),
\end{equation}
where $\bigoplus$ is a permutation-invariant aggregator and $\phi^{(\ell)}, \psi^{(\ell)}$ are learnable maps; stacking $L$ such layers yields a representation whose receptive field is the $L$-hop neighbourhood of each node.

From the \emph{graph signal processing} perspective, the same operation can be written as the application of a graph convolutional filter — in the sense of Equation~\eqref{eq:graph-filter} — to the matrix of node features $\mathbf{H}^{(\ell)} \in \mathbb{R}^{|V|\times F_\ell}$, followed by a pointwise non-linearity:
\begin{equation}
    \mathbf{H}^{(\ell+1)} \;=\; \sigma\bigl(\mathrm{\Phi}(\mathbf{S})\,\mathbf{H}^{(\ell)}\,\mathbf{W}^{(\ell)}\bigr),
\end{equation}
where $\mathbf{S}$ is a graph shift operator and $\mathbf{W}^{(\ell)}$ collects the learnable channel-mixing weights. The two viewpoints are equivalent: choosing an aggregator $\bigoplus$ in the message-passing formulation corresponds to choosing a shift operator $\mathbf{S}$ and a filter order $K$ in the spectral one, and the spectral response of $\mathrm{\Phi}(\mathbf{S})$ — encoded in the polynomial coefficients $\{\varphi_k\}_{k=0}^{K}$ — governs which frequencies of the input signal are amplified or attenuated under propagation \cite{ortega2018graph}. The two architectures we consider differ along precisely this axis. GCN uses a degree-normalised mean aggregation, which from the signal-processing side amounts to a fixed first-order polynomial in the symmetrically normalised adjacency matrix $\tilde{\mathbf{A}} = \mathbf{D}^{-1/2}\mathbf{A}\mathbf{D}^{-1/2}$ and acts as a low-pass filter that progressively smooths the graph signal across layers. GIN, by contrast, uses a sum aggregation followed by a multilayer perceptron and is provably as expressive as the Weisfeiler--Leman graph isomorphism test \cite{xu2018powerful}; in spectral terms, the sum aggregation corresponds to a filter built from the unnormalised adjacency $\mathbf{A}$, which preserves degree information that the GCN normalisation discards — a difference that turns out to matter substantially on our dataset.

Three design choices specific to our setting deserve mention. First, our dataset consists of pure graph structures with no node features. We therefore inject a trivial input signal — either a constant vector of ones or the first $k$ eigenvectors of the graph Laplacian — and let the GNN diffuse it across the topology in a manner analogous to heat diffusion on a graph \cite{kondor2002diffusion}; the resulting node embeddings depend only on the graph structure, which is the object of classification. Second, because the dataset spans Cayley graphs of widely varying group order — and hence size and diameter — the standard pathologies of message-passing interact non-trivially with model depth. \emph{Over-squashing} of long-range information through bottleneck edges \cite{alon2020bottleneck} is largely neutralised by the regularity and vertex-transitivity of Cayley graphs, but \emph{over-smoothing} — the convergence of node representations towards the kernel of the propagation operator under repeated low-pass filtering \cite{li2018deeper, oono2020graph} — remains a real constraint: a depth that is benign on large Cayley graphs may over-smooth the smaller ones, and the low-pass character of the GCN filter aggravates this regime. The order-256 experiment below partially controls for graph size, but does not by itself isolate the effect of model depth. Third, because the task is graph-level classification, the final layer of each GNN aggregates node embeddings via global mean pooling, yielding a size-normalised graph representation that avoids biasing the classifier towards features merely correlated with the group order $|G|$ \cite{grattarola2022understanding}.

\subsection{Results}
\label{sec:ML_results}

The results in Table~\ref{tab:all-models-summary-not256} show that the Cayley graphs contain a strong signal for most of the properties considered here, but that this signal is not equally accessible to all representations. The classical models are trained on precomputed graph statistics, whereas the MLP and the GNNs are trained directly on the Cayley graph. Thus, the comparison is not simply between different classifiers, but also between two different ways of presenting the same mathematical object to a learning algorithm.

For the tabular models on the non-order-256 subset, the tree-based methods are consistently stronger than logistic regression. Logistic regression often achieves a high ROC AUC, but its precision can be poor on the rarer positive classes. This is especially clear for the abelian and monolithic targets, where the model recovers many positive examples but at the cost of many false positives. Random Forest and XGBoost are better suited to this setting, since the graph statistics do not interact linearly with the labels. For the abelian target, XGBoost achieves the highest F1 score, while Random Forest gives the highest precision. For the nilpotent target, Random Forest gives the strongest tabular result. XGBoost is slightly more recall-oriented on this task, but with lower precision and F1. These results indicate that the hand-crafted graph invariants contain substantial information about nilpotency, but that extracting it requires nonlinear decision boundaries.

The monolithic task is harder. On the non-order-256 subset the positive rate is only $0.015$, and all three tabular models show the usual tension between recall and precision under strong class imbalance. Logistic regression obtains the highest recall, but with very low precision. Random Forest is much more conservative, giving the best precision and F1 score, while XGBoost again favours recall. The order-256 experiment reinforces this picture. Once the group order is fixed, the monolithic classification problem becomes more difficult, and the F1 scores drop substantially. This suggests that part of the signal available in the larger dataset is correlated with group order, while the fixed-order setting forces the models to rely on finer structural features of the Cayley graph.

Some targets are much easier. The tabular models classify cyclic groups perfectly on the non-order-256 subset, which is likely explained by the inclusion of the minimum number of generators as an input feature: a finite group is cyclic exactly when it is generated by one element. Solvability is also essentially solved by the ensemble methods. The simple and perfect tasks should be interpreted more cautiously, because the positive classes are extremely rare. In these cases, accuracy and ROC AUC alone can be misleading, and precision, recall, F1, and class balance have to be considered together. Nevertheless, the strong performance of the ensemble methods shows that the chosen network statistics are highly informative when the relevant positive examples are present.

The graph-model results give a different perspective. GIN performs very strongly across the non-256 dataset, and in several cases it reaches or exceeds the performance suggested by the tabular baselines. The most important example is the nilpotent target. In Table~\ref{tab:all-models-summary-not256}, the strongest tabular model is Random Forest. XGBoost gives slightly higher recall, but with lower precision and F1. In the same table, GIN improves on the tabular models across all four metrics. This makes nilpotency one of the clearest cases where the graph neural network is not only competitive with the tabular methods, but performs better. This is particularly interesting because the tabular methods are given access to precomputed graph statistics, including expensive features such as the minimal cycle basis, whereas GIN learns directly from the graph structure. The result suggests that the message-passing representation is able to recover structural information relevant to nilpotency without requiring these hand-crafted graph descriptors.

This point is important for the interpretation of the whole experiment. The tabular methods are strong baselines, but they are given a considerable amount of preprocessed information. Before training logistic regression, Random Forest, or XGBoost, one must compute graph-level quantities such as diameter, girth, clustering coefficients, distance-based indices, triangle counts, and the summary statistics of the minimal cycle basis. These quantities are mathematically meaningful, but they are not free. In particular, computing the minimal cycle basis is one of the more expensive steps in the preprocessing pipeline, since the dimension of the cycle space grows as $|E|-|V|+1$ for a connected graph and basis algorithms typically require repeated shortest-path or linear-algebra operations on graph-scale data. It therefore becomes a practical bottleneck as the size and number of Cayley graphs increase. The tabular models therefore shift part of the learning problem into feature engineering and feature computation.

The GNNs do not require this same preprocessing. They take the Cayley graph itself as input and build a graph-level representation by message passing and pooling. In this sense, GIN is learning from a less curated representation of the data: it is not given the cycle-basis statistics, distance indices, or clustering summaries explicitly. Its strong performance, especially on the nilpotent target, suggests that the relevant algebraic information is present in the graph structure in a form that can be recovered by a sufficiently expressive message-passing architecture. This does not make the tabular models redundant; rather, it shows that the two approaches answer slightly different questions. The tabular models show that known graph invariants are predictive of Cayley graph properties. GIN shows that some of the same, and in some cases stronger, predictive signal can be learned directly from the graph.

The comparison between GCN and GIN also supports this interpretation. GCN performs well on some of the highly separable or extremely imbalanced tasks, including the simple, perfect, and solvable targets, where its accuracy, macro precision, and MCC are high. This indicates that even the simpler normalised message-passing architecture can detect some coarse structural distinctions in the Cayley graphs. However, its behaviour is much less stable than GIN. On the abelian, nilpotent, monolithic, and cyclic targets, the GCN results are poor, and the very large losses suggest severe calibration or optimisation difficulties. This is consistent with the discussion above: GCN applies a degree-normalised, low-pass aggregation, and on regular Cayley graphs this can quickly smooth away useful distinctions between node representations. GIN, using sum aggregation followed by an MLP, is more expressive and is better able to preserve the structural information needed for graph-level classification.

We also compare the MLP against the other models. 
The MLP, trained directly on the Cayley graph rather than on precomputed statistics, gives a mixed picture relative to the tabular and graph-based models. On several targets its recall sits almost exactly at 0.500 (abelian, nilpotent, monolithic, and cyclic), which is a strong indication that the model is not learning a discriminative decision boundary for these tasks and is instead defaulting to something close to chance-level separation between classes.
This is likely a consequence of the severe class imbalance combined with a fixed 0.5 probability threshold. 
In contrast with GIN, which is trained on the same raw graph input but achieves much higher recall on the same targets, this suggests that the gap is attributable to architecture rather than to the input representation alone: sum-based message passing followed by pooling appears far better suited to extracting graph-level structural signal than a feedforward network applied without explicit relational structure.

The clearest exception is the perfect target, where the MLP is the only model to recover any positive examples at all: Random Forest and XGBoost both collapse to predicting the negative class entirely (precision, recall, and F1 all 0.000), while the MLP achieves precision 0.996 and recall 0.435.
This suggests that, despite its weaker performance elsewhere, the MLP's smoother decision function is more robust than the tree-based models to extreme class rarity, where a tree-based model can trivially satisfy its splitting criteria by never predicting the minority class.

Overall, the experiments suggest that Cayley graphs encode a substantial amount of information about the properties of the corresponding groups. When informative graph statistics are precomputed, classical ensemble methods perform very well and provide strong baselines. However, this strength depends on a potentially expensive feature pipeline, with the minimal cycle basis being a particularly costly component. GIN gives a complementary result: it shows that a graph neural network can recover much of this information directly from the Cayley graph, and for nilpotency it appears to give the strongest performance among the approaches tested. The broader conclusion is therefore not that one representation uniformly dominates the other, but that graph neural networks offer a promising route when one wants to avoid hand-crafted graph descriptors, or when the relevant descriptors are unknown or expensive to compute.

The results in Table~\ref{tab:ord256_results} for the order-256 subset gives a more cautious picture. In this experiment the only target considered is monolithic, as other targets did not have a meaningful split (as discussed in Section~\ref{sec:order-256-analysis}), and the positive rate is just $0.020$. Logistic regression obtains the highest recall, but its precision is lower, so it identifies many of the positive examples at the cost of a large number of false positives. Random Forest has the opposite behaviour: it obtains the joint-best accuracy, but its recall is lower, meaning that it misses most of the positive cases.

The strongest overall results are split between XGBoost, GIN, and GCN. XGBoost and GIN share the best ROC AUC, while GIN also shares the best accuracy with Random Forest. GCN gives the highest precision, recall, and F1 score, although its ROC AUC is not reported in this comparison. This suggests that the graph neural networks are competitive on the order-256 subset, but these results are not enough to conclude that they are definitively better than the tabular methods in this fixed-order setting.

The low positive rate is likely a major reason why the task is difficult for all models. With so few positive examples, the learned decision boundary is sensitive to the particular train-test split, and small changes in the number of true positives can have a large effect on precision, recall, and F1. The fixed-order setting may also remove some of the easier signal that is available in the broader dataset, forcing the models to distinguish monolithic groups using more subtle structural properties of the Cayley graphs. A useful next step would be to repeat this experiment over several stratified splits, report confidence intervals, and consider threshold tuning or class-weighted training, especially for GIN and XGBoost. This would help determine whether the apparent differences between the models are robust, or whether they are mainly a consequence of the severe class imbalance.

\begin{table}[H]
\centering
\small
\caption{Classification results for the non-order 256 subset. For each target property the best value among models is bolded within each metric. Positive rate is the frequency of the property in the dataset.}
\label{tab:all-models-summary-not256}
\begin{tabular}{llrrrrr}
\toprule
Target & Model & Pos. rate & ROC AUC & Prec. & Rec. & F1 \\
\midrule
\multirow[t]{6}{*}{Abelian} & LogReg & \multirow[t]{6}{*}{0.020} & 0.989 & 0.393 & 0.947 & 0.556 \\
 & RF &  & 0.999 & \textbf{0.955} & 0.773 & 0.855 \\
 & XGB &  & \textbf{1.000} & 0.882 & 0.987 & \textbf{0.932} \\
 & MLP & & - & 0.628 & 0.496 & 0.555\\
 & GCN &  & 0.500 & 0.019 & \textbf{1.000} & 0.038 \\
 & GIN &  & 0.999 & 0.877 & 0.934 & 0.905 \\
\midrule
\multirow[t]{6}{*}{Nilpotent} & LogReg & \multirow[t]{6}{*}{0.160} & 0.900 & 0.511 & 0.768 & 0.614 \\
 & RF &  & 0.997 & 0.959 & 0.908 & 0.933 \\
 & XGB &  & 0.996 & 0.844 & 0.967 & 0.901 \\
 & MLP & & - & 0.422 & 0.500 & 0.458\\
 & GCN &  & 0.500 & 0.161 & \textbf{1.000} & 0.278 \\
 & GIN &  & \textbf{1.000} & \textbf{0.986} & 0.995 & \textbf{0.990} \\
\midrule
\multirow[t]{6}{*}{Simple} & LogReg & \multirow[t]{6}{*}{0.002} & \textbf{1.000} & 0.493 & 0.971 & 0.654 \\
 & RF &  & 0.971 & \textbf{1.000} & 0.943 & 0.971 \\
 & XGB &  & \textbf{1.000} & 0.971 & 0.943 & 0.957 \\
 & MLP & & - & 0.940 & 0.509 & 0.661\\
 & GCN &  & 0.500 & 0.000 & 0.000 & 0.000 \\
 & GIN &  & \textbf{1.000} & \textbf{1.000} & \textbf{1.000} & \textbf{1.000} \\
\midrule
\multirow[t]{6}{*}{Perfect} & LogReg & \multirow[t]{6}{*}{0.000} & \textbf{1.000} & 0.038 & \textbf{1.000} & 0.074 \\
 & RF &  & 0.749 & 0.000 & 0.000 & 0.000 \\
 & XGB &  & 0.997 & 0.000 & 0.000 & 0.000 \\
 & MLP & & - & \textbf{0.996} & 0.435 & \textbf{0.605}\\
 & GCN &  & 0.500 & 0.000 & 0.000 & 0.000 \\
 & GIN &  & 0.966 & 0.000 & 0.000 & 0.000 \\
\midrule
\multirow[t]{6}{*}{Solvable} & LogReg & \multirow[t]{6}{*}{0.999} & \textbf{1.000} & \textbf{1.000} & 0.995 & 0.998 \\
 & RF &  & \textbf{1.000} & \textbf{1.000} & \textbf{1.000} & \textbf{1.000} \\
 & XGB &  & \textbf{1.000} & \textbf{1.000} & 0.999 & \textbf{1.000} \\
 & MLP & & - & 0.963 & 0.999 & 0.999\\
 & GCN &  & 0.500 & 0.999 & \textbf{1.000} & 0.999 \\
 & GIN &  & 0.992 & 0.999 & \textbf{1.000} & 0.999 \\
\midrule
\multirow[t]{6}{*}{Monolithic} & LogReg & \multirow[t]{6}{*}{0.015} & 0.921 & 0.080 & 0.831 & 0.146 \\
 & RF &  & 0.984 & 0.791 & 0.476 & 0.595 \\
 & XGB &  & \textbf{0.986} & 0.396 & 0.810 & 0.532 \\
 & MLP & & - & \textbf{0.879} & 0.500 & \textbf{0.637}\\
 & GCN &  & 0.500 & 0.015 & \textbf{1.000} & 0.030 \\
 & GIN &  & 0.933 & 0.446 & 0.330 & 0.380 \\
\midrule
\multirow[t]{6}{*}{Cyclic} & LogReg & \multirow[t]{6}{*}{0.010} & \textbf{1.000} & \textbf{1.000} & \textbf{1.000} & \textbf{1.000} \\
 & RF &  & \textbf{1.000} & \textbf{1.000} & \textbf{1.000} & \textbf{1.000} \\
 & XGB &  & \textbf{1.000} & \textbf{1.000} & \textbf{1.000} & \textbf{1.000} \\
 & MLP & & - & 0.758 & 0.500 & 0.602\\
 & GCN &  & 0.500 & 0.010 & \textbf{1.000} & 0.019 \\
 & GIN &  & 0.982 & 0.664 & 0.529 & 0.589 \\
\bottomrule
\end{tabular}
\end{table}

\begin{table}[H]
\centering
\caption{Classification results for the order-256 subset. The best value among models is bolded within each metric.}
\label{tab:all-models-summary-order-256}
\begin{tabular}{llrrrrrr}
\toprule
Target & Model & Pos. rate & Acc. & ROC AUC & Prec. & Rec. & F1 \\
\midrule
\multirow[t]{6}{*}{Monolithic}
& LogReg & \multirow[t]{6}{*}{0.020} & 0.851 & 0.930 & 0.105 & 0.854 & 0.187 \\
& RF     &                            & \textbf{0.981} & 0.950 & 0.552 & 0.212 & 0.307 \\
& XGB    &                            & 0.954 & \textbf{0.959} & 0.260 & 0.699 & 0.379 \\
& MLP & & 0.980 & - & 0.627 & 0.497 & 0.554\\
& GCN    &                            & 0.979 & - & \textbf{0.853} & \textbf{0.863} & \textbf{0.858} \\
& GIN    &                            & \textbf{0.981} & \textbf{0.959} & 0.569 & 0.274 & 0.370 \\
\bottomrule
\end{tabular}
\label{tab:ord256_results}
\end{table}

%%%%%%%%%%%%%%%%%%%%%%%%%%%%%%%%%%%%%%%%%%%%%%%%%
\section{Summary}\label{sec:summary}
This work develops a large-scale computational study of finite groups through the network geometry of their Cayley graphs. We construct a dataset containing $131{,}406$ Cayley graphs, one for each group of order at most $767$ excluding order $512$, together with group properties, and an array of network properties. The resulting dataset is both comprehensive within this range and practically useful: it links exact group-theoretic properties from GAP with computable network observables, making it a structured benchmark for testing which algebraic features are visible from Cayley graph data and for comparing classical and graph-based learning methods on mathematically labelled graphs.

A second contribution is the extraction of new enumerative information from this census. The counts of cyclic, abelian, nilpotent, simple, perfect, and solvable groups recover known OEIS sequences where such references exist, providing a consistency check on the data generation pipeline. Beyond this validation, the work contributes new OEIS sequences for the number of monolithic groups, and for the numbers of groups generated by at most three, four, and five elements. These sequences give reusable reference data for future computational group theory and illustrate how the dataset can produce mathematical outputs beyond the following machine-learning experiments.

The network analysis also identifies several empirical regularities that suggest new mathematical conjectures. These include the vanishing of directed square clustering for perfect groups under suitable generating-set restrictions, sharp behaviour of Cayley graph diameter relative to group order and generator number, an unexpectedly tight relationship between average graph disorder and order-scaled diameter, and a spectral conjecture linking large eigengaps in nilpotent Cayley graphs to quotients in the upper central series. They are motivated by broad computational evidence but remain open to proof, refinement, or counterexample through further algebraic and spectral analysis.

The machine-learning results show that Cayley graphs contain substantial predictive signal for several structural group properties. Classical ensemble methods perform strongly when supplied with precomputed graph statistics, confirming that these invariants encode meaningful algebraic information. At the same time, the graph neural network results demonstrate the value of learning directly from the Cayley graph. In particular, GIN gives the strongest performance on nilpotency in the non-order-256 dataset, improving over the tabular baselines while avoiding hand-crafted descriptors such as minimal-cycle-basis statistics. The fixed-order experiment on monolithic groups gives a complementary case in which GNNs remain competitive after the easiest order-dependent signal is removed, with GCN achieving the best precision, recall, and F1 score in that setting. Overall, the comparison suggests that GNN architectures are especially useful when informative graph descriptors are expensive to compute, unknown in advance, or when the task depends on relational structure not captured by simple vector summaries.

Natural extensions include enlarging the census to higher orders and additional generating-set choices, separating generator-dependent effects from group-intrinsic phenomena, strengthening the conjectures with targeted computations on specific families, and repeating the learning experiments across multiple stratified splits with calibrated thresholds and uncertainty estimates. Further work could also incorporate richer node or edge features, undirected variants of selected statistics, and spectral or representation-theoretic features designed specifically for Cayley graphs.

Taken together, the results show that Cayley graphs provide a productive bridge between finite group theory, network science, and graph machine learning, yielding a dataset, new sequences, conjectural mathematics, and predictive baselines that can support a wider programme of data-driven discovery in algebra.

%%%%%%%%%%%%%%%%%%%%%%%%%%%%%%%%%%%%%%%%%%%%%%%%%
\section*{Acknowledgements}
YH is supported by the Wallenberg AI, Autonomous Systems and Software Program (WASP) funded by the Knut and Alice Wallenberg Foundation.
EH is supported by São Paulo Research Foundation (FAPESP) grant 2024/18994-7. 
AO is supported by The Mark Foundation for Cancer Research Aspire Award grant 24-023-ASP. 
HS is supported by the Engineering and Physical Sciences Research Council [EP/S021590/1], the EPSRC Centre for Doctoral Training in Geometry and Number Theory (The London School of Geometry and Number Theory), King's College London. 
The authors thank the organisers of the LOGML25 school, where this project was initiated.

%%%%%%%%%%%%%%%%%%%%%%%%%%%%%%%%%%%%%%%%%%%%%%%%%
%\newpage
\addcontentsline{toc}{section}{References}
\bibliographystyle{utphys}
\bibliography{references}{}

\end{document}